%% file: main.tex
\documentclass[10pt,journal,compsoc]{IEEEtran}

\usepackage{times}
\usepackage{epsfig}
\usepackage{graphicx}
\usepackage{amsmath}
\usepackage{amssymb}
\usepackage{subfig}
\usepackage{float}
\usepackage{xcolor,colortbl}
\usepackage{multirow}
\usepackage{booktabs}

\usepackage[breaklinks=true,bookmarks=false]{hyperref}
\usepackage{review}

\setrevision{0}

\newcommand{\comRev}[1]{#1}
\newcommand{\comRevSec}[1]{{#1}}
\definecolor{darkgreen}{gray}{0.0}

\ifCLASSOPTIONcompsoc
  \usepackage[nocompress]{cite}
\else
  \usepackage{cite}
\fi

\ifCLASSINFOpdf
\else
\fi
\begin{document}
%
\title{Appearance and Pose-Conditioned Human Image Generation using Deformable GANs}
%
%
%
%

\author{Aliaksandr~Siarohin,
        St{\'e}phane~Lathuili{\`e}re,
        Enver~Sangineto
        and~Nicu~Sebe
\IEEEcompsocitemizethanks{\IEEEcompsocthanksitem Aliaksandr Siarohin, Enver Sangineto, 
and Nicu Sebe are with the Department of Information Engineering and Computer Science (DISI),
University of Trento, Italy.\protect\\
E-mail: aliaksandr.siarohin@unitn.it; enver.sangineto@unitn.it; sebe@disi.unitn.it
}
\IEEEcompsocitemizethanks{\IEEEcompsocthanksitem St{\'e}phane Lathuili{\`e}re is with LTCI, T\'{e}l\'{e}com Paris, Institut polytechnique de Paris, France.\protect\\
E-mail:  stephane.lathuiliere@telecom-paris.fr
}
}

%
%

\markboth{Journal of \LaTeX\ Class Files,~Vol.~14, No.~8, August~2015}%
{Shell \MakeLowercase{\textit{et al.}}: Deformable GAN}
%



\IEEEtitleabstractindextext{%
\begin{abstract}
In this paper, we address the problem of generating person images conditioned on both pose and appearance information. 
Specifically, given an image $x_a$ of a person and a target pose $P(x_b)$, extracted from an image $x_b$, we synthesize a new image of that person  in  pose $P(x_b)$, while
preserving the visual details in $x_a$. In order to deal with pixel-to-pixel misalignments caused by the pose differences between $P(x_a)$ and $P(x_b)$, we introduce {\em deformable skip connections} in  the  generator of our Generative Adversarial Network. Moreover, a {\em nearest-neighbour loss} is proposed instead of the common $L_1$ and $L_2$ losses in order to match the details of the generated image with the target image.
Quantitative and qualitative results, using common datasets and protocols recently proposed for this task, show that our approach is competitive with respect to the state of the art. Moreover, we conduct an extensive evaluation using 
off-the-shell person re-identification (Re-ID) systems trained with person-generation based  augmented data, which is one of the main important applications for this task. Our experiments show that our Deformable GANs can significantly boost the Re-ID accuracy and are even better than data-augmentation methods  specifically trained using Re-ID losses. 
\end{abstract}

}

\maketitle

\IEEEdisplaynontitleabstractindextext

%
\IEEEpeerreviewmaketitle

\input{intro}
\input{related}

\input{model}
\input{implementation}

\input{experimentation}

\section{Conclusions}
\label{sec:ccl}
In this paper we presented a GAN-based approach for image generation of persons conditioned on the appearance and  the pose. We introduced two novelties: deformable skip connections and nearest-neighbour loss. The first are used to solve common problems in U-Net based generators when dealing with deformable objects. 
The second is used to alleviate a different type of misalignment between the generated image and the ground-truth image.

Our experiments, based on both automatic evaluation metrics and human judgements,
show that the proposed method  outperforms 
or is comparable with 
 previous work on this task.
Importantly, we show that, contrary to other {\em generic} person-generation methods, our Deformable GANs can be used to significantly improve the accuracy of different Re-ID systems using 
 data-augmentation and the obtained performance boost is even higher than a state-of-the-art Re-ID {\em specific} data-augmentation approach.
 
Despite we tested our  Deformable GANs on the specific task of human-generation, only few assumptions are used which refer to the human body and we believe that our proposal can be easily extended to address other deformable-object generation tasks.

\ifCLASSOPTIONcompsoc
  \section*{Acknowledgments}
\else
  \section*{Acknowledgment}
\fi

We want to thank the NVIDIA Corporation for the  donation  of  the  GPUs  used  in this project.\\
This project has received funding from the European Research Council (ERC) (Grant agreement No.788793-BACKUP).

\ifCLASSOPTIONcaptionsoff
  \newpage
\fi



%

\bibliographystyle{IEEEtran}
\bibliography{egbib}

%

\begin{IEEEbiography}[{\includegraphics[width=1in,height=1.25in,clip,keepaspectratio]{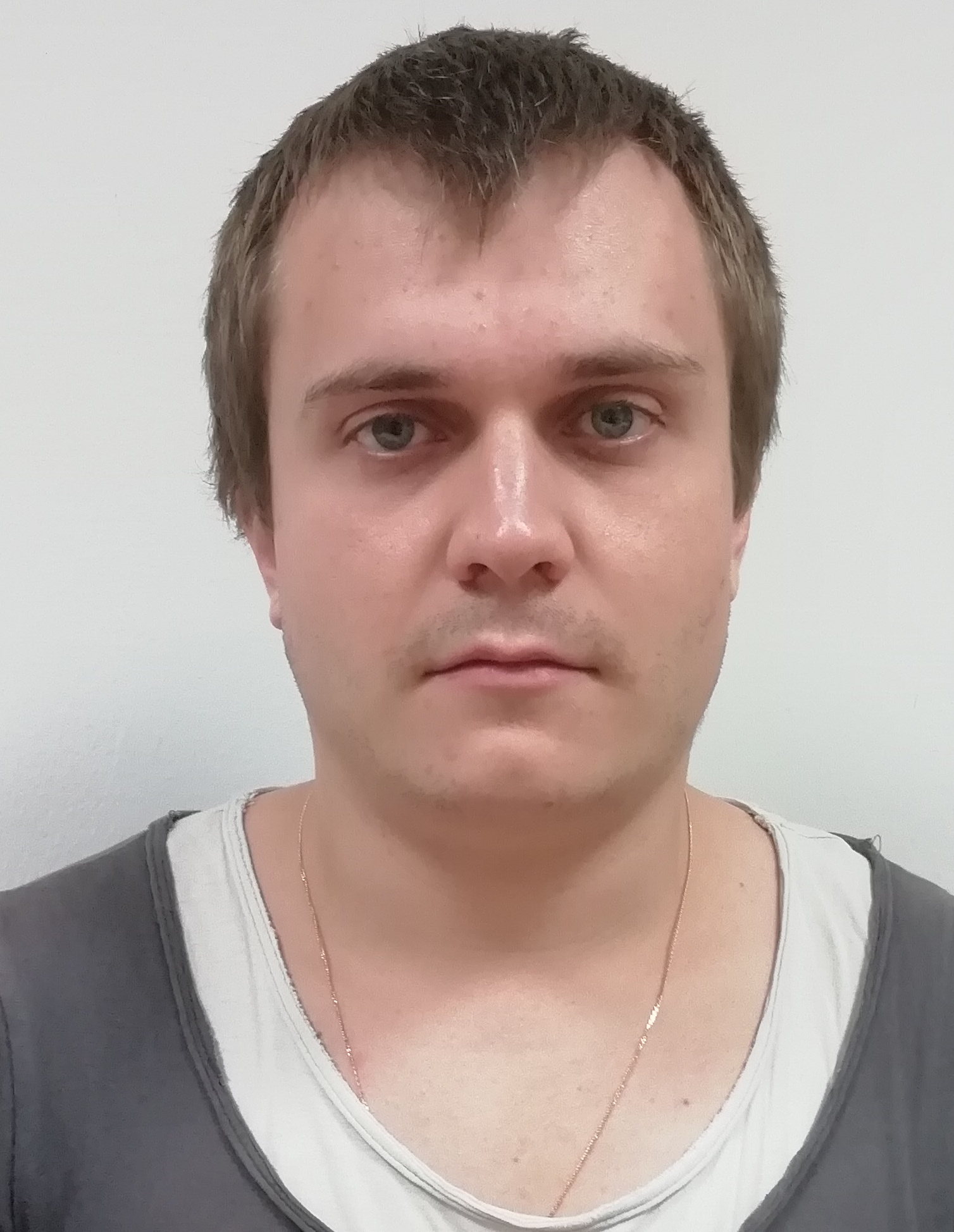}}]{Aliaksandr Siarohin}
received the B.Sc. degree in Applied Mathematics from Belarussin State University in 2015.
He worked towards the M.Sc. degree in computer science and obtained it from University of Trento in 2017. He is currently a Phd candidate at
the University of Trento. His research interests include machine learning for image and video generation, generative adversarial networks and domain adaptation.
\end{IEEEbiography}

\begin{IEEEbiography}[{\includegraphics[width=1in,height=1.25in,clip,keepaspectratio]{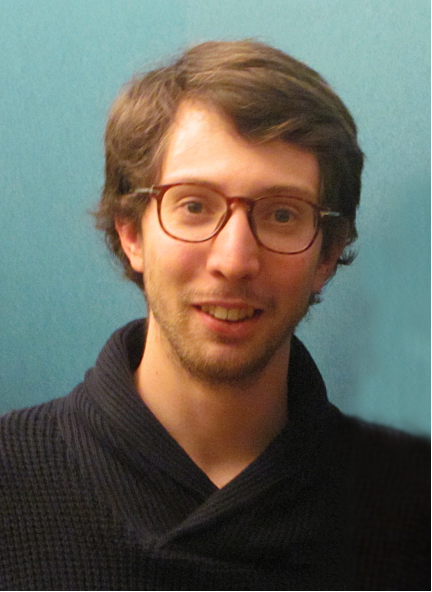}}]{St\'{e}phane Lathuili\`{e}re} is an assistant professor
at T\'{e}l\'{e}com ParisTech, France. He received the
M.Sc. degree in applied mathematics and computer science from Grenoble Institute of Technology in 2014. He worked towards his Ph.D. in the Perception Team at Inria under the supervision of Dr. Radu Horaud, and obtained it from Universit\'{e} Grenoble Alpes (France) in 2018. He was a Postdoc researcher at the Univeristy of Trento.
His research interests cover deep models for regression, generation and domain adaptation.
\end{IEEEbiography}

\begin{IEEEbiography}[{\includegraphics[width=1in,height=1.25in,clip,keepaspectratio]{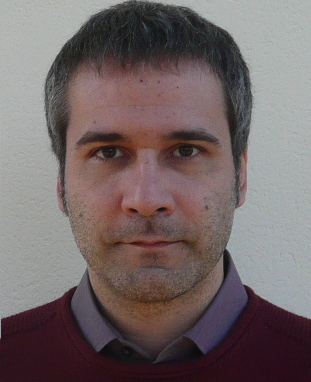}}]{Enver Sangineto}
is   Assistant Professor at University of Trento, Department of Information Engineering and Computer Science. 
He received his PhD in Computer Engineering from the University of Rome ``La Sapienza''.
After that he has been a post-doctoral researcher at 
the Universities of Rome
``Roma Tre'' and ``La Sapienza'' and at the Italian Institute of Technology (IIT) in Genova.
 His research interests include generative methods and 
learning with minimal human supervision.
\end{IEEEbiography}

\begin{IEEEbiography}[{\includegraphics[width=1in,height=1.25in,clip,keepaspectratio]{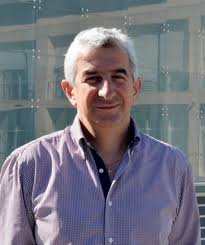}}]{Nicu Sebe}
is Professor with the University of
Trento, Italy, leading the research in the areas
of multimedia information retrieval and human
behavior understanding. He was the General
Chair of the FG 2008 and
ACM Multimedia 2013, and the Program Chair
of the International Conference on Image and
Video Retrieval in 2007 and 2010, ACM Multimedia
2007 and 2011. He was the Program Chair
of ICCV 2017 and ECCV 2016, and a General
Chair of ACM ICMR 2017. He is a fellow of IAPR.
\end{IEEEbiography}



\end{document}

%% file: intro.tex
\section{Introduction}
\label{Introduction}
The appearance and pose-conditioned human image generation  task aims at generating a person image  conditioned on two different variables: (1) the appearance of a specific person in a given image and (2) the pose 
of the same 
person in another image.
Specifically, the generation process needs to preserve the appearance details (e.g., colors of the clothes, texture, etc.) contained in the first variable while performing a deformation on the structure (the pose) of the foreground person according to the second variable.
Generally speaking, this task can be extended to the generation of images where the foreground, i.e., a deformable object such as a face or a body, changes because of a viewpoint variation or a deformable motion. The common assumption is that the object structure can be automatically extracted using a keypoint detector.

\begin{figure}[t!]
\centering
\subfloat[Aligned task]{\includegraphics[height=0.45\columnwidth]{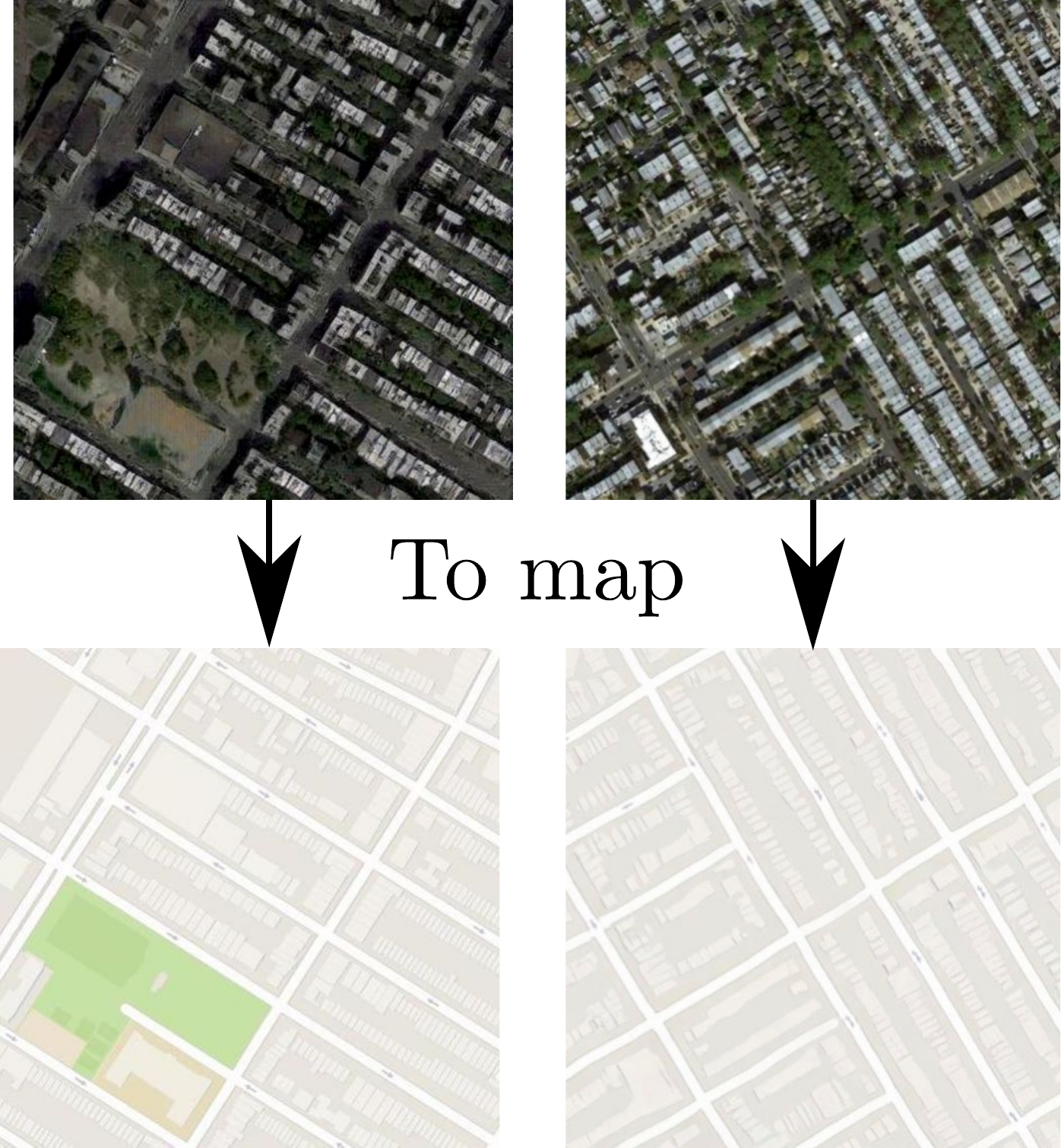}\label{fig:teaserAligned}}
\hspace{0.5cm}
\subfloat[Unaligned task]{\includegraphics[height=0.45\columnwidth]{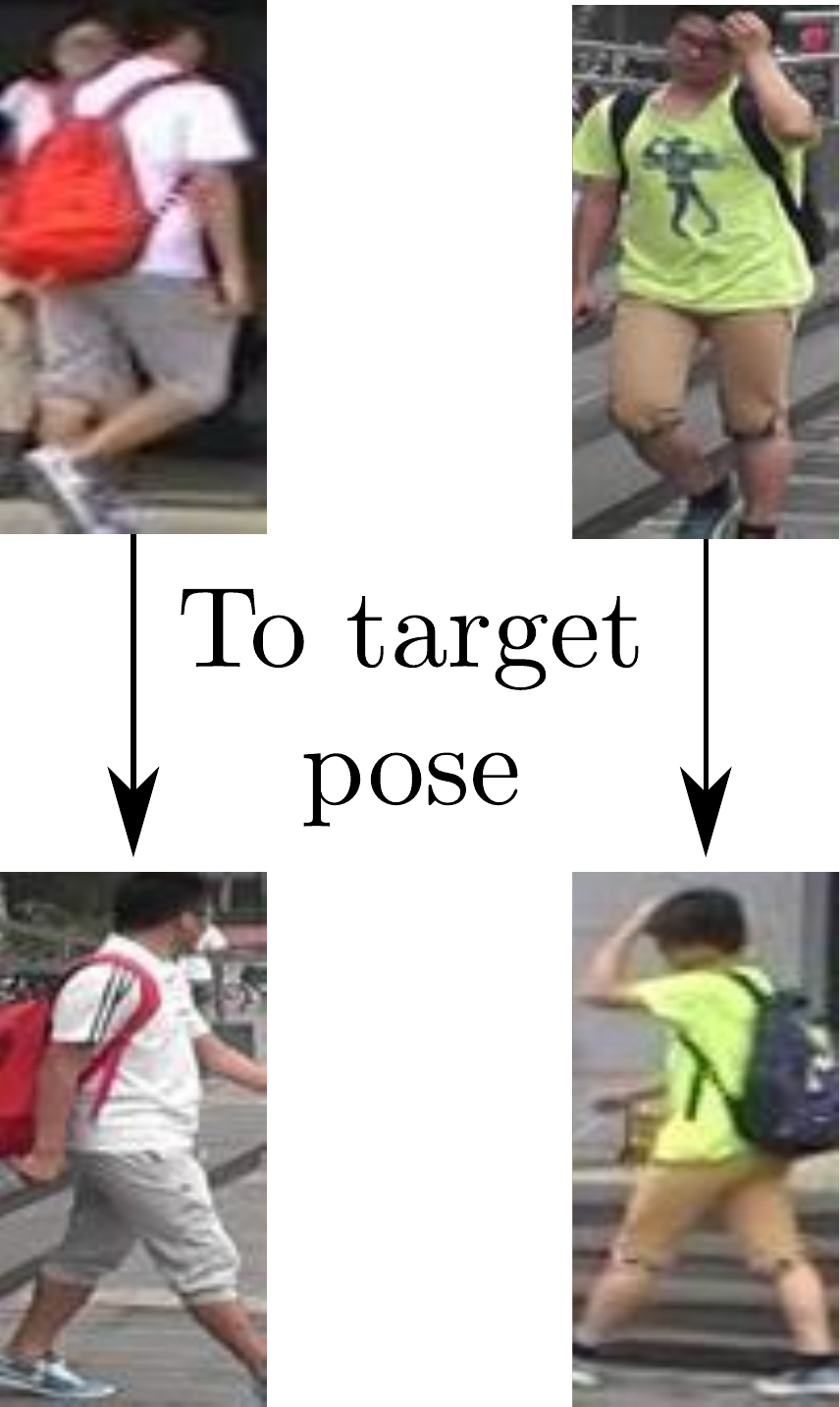}\label{fig:teaserUnaligned}}
\caption{(a) An example of a ``rigid'' scene generation task, where the conditioning and the output image local structures are well aligned. (b) In a deformable-object  generation task, the input and output images are not spatially aligned.}
\vspace{-0.5cm}
\label{fig:teaser}
\end{figure}

After the publication of the pioneering work of Ma et al. \cite{ma2017pose}, there has been a quickly growing interest in this task, as witnessed by some very recent papers on this topic \cite{walker2017pose,ZhaoWCLF17,esser2018variational,si2018multistage,liu2018pose,siarohin2018deformable}.
The reason of this large interest is probably due to the many potential applicative scenarios, ranging from 
 computer-graphics based manipulations \cite{walker2017pose}
 to data augmentation for training
    person re-identification (Re-ID)  \cite{Zheng_2017_ICCV,liu2018pose}
   or  human pose estimation \cite{Cao} systems. 
However, most of the recently proposed, deep-network based generative approaches, such as  Generative Adversarial Networks (GANs) \cite{goodfellow2014generative} or Variational Autoencoders (VAEs) 
 \cite{kingma2013auto} do not explicitly deal with the problem of articulated-object generation.
 Common conditional methods (e.g., conditional GANs or conditional VAEs) can synthesize images whose appearances depend on some conditioning variables (e.g., a label or another image). For instance, Isola et al. \cite{pix2pix2016} proposed an ``image-to-image translation'' framework, in which an input image $x$ is transformed into a second image $y$ represented in another ``channel'' (see Fig.~\ref{fig:teaserAligned}).
However, most of these methods have problems when dealing with large spatial deformations between the conditioning appearance and the target image. 
For instance, 
the U-Net architecture used by  Isola et al. \cite{pix2pix2016} is based on {\em skip connections} which help preserving local information between $x$ and $y$. 
Specifically, skip connections are used to copy and then concatenate the feature maps of the generator ``encoder'' (where information is downsampled using convolutional layers) to the generator ``decoder'' (containing the upconvolutional layers).
However, the assumption used in \cite{pix2pix2016} is that $x$ and $y$ are roughly aligned with each other and they represent the same underlying structure. This assumption is violated 
when the foreground object in $y$ undergoes large spatial deformations with respect to $x$ (see Fig.~\ref{fig:teaserUnaligned}). \addnote[face]{2}{By opposition to human face~\cite{Sun_2018,Wang_2018}, human body is a highly non-rigid object and the important misalignment between the input and output poses may impact the generated image quality.}~As shown in \cite{ma2017pose}, skip connections cannot reliably cope with misalignments between the two poses. In Sec.~\ref{Related} we will see that U-Net based generators are widely used in 
most of the recent person-generation approaches, hence this misalignment problem is orthogonal to many methods.

Ma et al. \cite{ma2017pose} propose to alleviate this problem by using a two-stage generation approach. In the first stage, a U-Net generator is trained using a masked $L_1$ loss in order to produce an intermediate image conditioned on the target pose. In the second stage, a second U-Net based generator is trained using also an adversarial loss in order to generate an appearance difference map which brings the  intermediate image closer to the appearance of the conditioning image.
In contrast, the U-Net-based method we propose in this paper  is end-to-end trained
  by explicitly taking into account  pose-related spatial deformations. More specifically, we propose {\em deformable skip connections} which ``move'' local information according to the structural deformations represented in the  conditioning variables. These layers are used  in our U-Net based generator. 
In order to  move information according to specific spatial deformations, first we decompose the overall  deformation by means of a set of 
local affine  transformations involving subsets of joints. 
After that,  we deform the convolutional feature maps of the encoder  according to these  transformations and we use common skip connections to transfer the transformed tensors to the decoder's fusion layers.
Moreover, we also propose using a {\em nearest-neighbour 
  loss} as a replacement of common pixel-to-pixel losses (e.g., $L_1$ or $L_2$ losses) commonly used in conditional generative approaches. This loss proved to be helpful in generating local details (e.g., texture) similar to the target image which are not penalized because of small spatial misalignments.

  Part of the material presented here appeared in \cite{siarohin2018deformable}.
  The current paper 
 extends \cite{siarohin2018deformable} in
several ways. 
First, we present a more detailed  analysis of  related work by including very recently  published papers
dealing with pose-conditioned human image generation.
Second, we show how a variant of our method can be used in order to introduce a third conditioning variable: the background, represented by a third input image. 
Third, we describe in more details our method.
Finally, we extend the quantitative and qualitative experiments by comparing our Deformable GANs with the very recent work in this area. Specifically, this comparison with the state of the art is performed using: (1) the protocols proposed by Ma et al. \cite{ma2017pose} and (2)
Re-ID based experiments.
The latter are motivated by the recent trend  of using generative methods for data-augmentation
\cite{walker2017pose,liu2018pose,DBLP:conf/iccv/HariharanG17,DBLP:journals/corr/abs-1712-01381,DBLP:journals/corr/abs-1712-00981,DBLP:journals/corr/abs-1801-05401}, 
and show that Deformable GANs can largely improve the accuracy of different Re-ID systems. Conversely, most of the other state-of-the-art methods generate new training samples which are {\em harmful} for 
Re-ID systems, leading to a significantly {\em worse} performance with respect to a non-augmented training dataset. Although tested on the specific human-body problem, our approach makes few human-related assumptions and can be easily extended to other domains involving the generation of  highly deformable objects. 
Our code   and our trained models
are publicly available\footnote{\url{https://github.com/AliaksandrSiarohin/pose-gan}}.


%% file: related.tex
\section{Related work}
\label{Related}

Most common deep-network-based
 approaches for visual content generation can be categorized as either Variational Autoencoders (VAEs) \cite{kingma2013auto} or Generative Adversarial Networks (GANs) \cite{goodfellow2014generative}. VAEs  are based on probabilistic graphical models and are trained by maximizing a lower bound of the corresponding data likelihood. GANs are based on two networks, a generator and a discriminator, which are trained simultaneously such that the generator tries to ``fool'' the discriminator and the discriminator learns how to distinguish between real and fake images.

 Isola et al. \cite{pix2pix2016} propose a  conditional GAN framework for image-to-image translation problems,
 where a given scene representation is ``translated'' into another representation.
The main assumption behind this framework is that there exists a spatial correspondence between the low-level information of the conditioning and the output image. VAEs and GANs are combined in \cite{ZhaoWCLF17} to generate realistic-looking multi-view images of clothes  from  a single-view
input image. The target view is 
fed
 to the model using a viewpoint label such as \emph{front} or \emph{left side} and a two-stage approach (pose integration and image refinement) is adopted. 
Ma et al. \cite{ma2017pose} propose  a more general approach  which allows synthesizing person images in any arbitrary pose. Similarly to our proposal, the input of their  model is a conditioning appearance image of the person and a target new pose defined by 18 joint locations. The target pose is described by means of binary maps where small circles represent the joint locations. This work has been extended  in \cite{ma2018disentangled} by learning disentangled representations of person images. More precisely, in the  generator, the pose, the foreground and background are separately encoded in order to obtain a disentangled description of the image. The input image is then reconstructed by combining the three descriptors. The major advantage of this approach is that it does not require pairs of images of the same person at training time. However, the generated images consequently suffer  from a lower level of realism.

Inspired by Ma et al. \cite{ma2017pose}, several methods have been recently proposed to generate human images. In \cite{LassnerPG17,ZhaoWCLF17},
the generation process is split into two different stages: pose generation and texture refinement. 
Si et al.~\cite{si2018multistage} propose multistage adversarial losses for generating images of a person in the same pose but from another {\em camera}  viewpoint. Specifically, the first generation stage generates the body pose in the new viewpoint. The second and the third stages generate  the foreground (i.e., the person) and the background, respectively.
Similarly to our proposal, Balakrishnan et al. \cite{balakrishnansynthesizing} partition the human body into different parts and separately deform each of them.
Their method is based on producing a set of segmentation masks, one per body part, plus a whole-body mask which separates the human figure from the background. However, in order for the model
 to segment the human figure  without relying on pixel-level annotations, training is based on pairs of conditioning images with the same background (e.g., frame pairs extracted from 
 the same video 
  with a static camera and background). This constraint prevents the use of this method in applications such as Re-ID data augmentation  in which  training images are usually taken in different environments.
 In contrast to \cite{LassnerPG17,ZhaoWCLF17,si2018multistage,balakrishnansynthesizing}, in this paper we show that a single-stage approach, trained end-to-end, can be used for the same task 
 obtaining higher qualitative results and that our method can be easily used as a useful black-box for Re-ID data augmentation.
 
Recently, Neverova et al. \cite{Neverova_2018_ECCV} propose to 
synthesize a  new image of the input person by blending different  
generated
 texture maps. 
This method is based on a  
 dense-pose estimation system \cite{Guler2018DensePose} which  maps pixels from  images to a common
surface-based coordinate framework.  
However, since the dense-pose estimator needs to be trained with a large-scale ground-truth dataset with image-to-surface correspondences manually annotated, 
\cite{Neverova_2018_ECCV} is not directly comparable with most of the other works (ours included) which rely on (sparse) keypoint detectors, whose training is based on a lower level of human supervision.

In \cite{esser2018variational}  a VAE is used to represent the  appearance and pose with two separated encoders. The appearance and pose descriptors are then
  concatenated and passed to a decoder  which generates the final image. 
Zanfir et al. \cite{zanfir2018human} estimate the human 3D-pose  
 using meshes, identifying the mesh regions which can be
transferred directly from the input image mesh to the target mesh. 
Finally, the missing surfaces are filled using a color regressor trained via Euclidean loss minimization.
Despite the visually satisfying results, this method requires prior knowledge in order to obtain the 3D body meshes and the 
clothes 
segmentation used to synthesize the final image.   
In \cite{liu2018pose}, a person generation model is specifically designed for boosting Re-ID accuracy using data augmentation. A sub-network is added to a standard U-Net GAN network in order to verify whether the identity of the person in the generated images can be distinguished from other identities.

Generally speaking, U-Net based architectures are frequently adopted for
pose-based person-image generation tasks \cite{LassnerPG17,ma2017pose,walker2017pose,ZhaoWCLF17,esser2018variational,si2018multistage,liu2018pose}. However, common U-Net skip connections  are not well-designed for large spatial deformations  because local information in the input and in the output images is not  aligned (Fig.~\ref{fig:teaser}b).
In contrast, we propose deformable skip connections to deal with this misalignment problem and ``shuttle'' local information from the encoder to the decoder driven by the specific pose difference.
In this way, differently from previous work, we are able to simultaneously  generate
the overall pose and the texture-level refinement.
Note that many of the above mentioned U-Net based methods are partially
 complementary to our approach since our deformable skip connections can potentially be plugged into the corresponding U-net, possibly increasing the final performance.

\addnote[landmarkGen]{1}{Landmark locations are exploited for other generation tasks such as face synthesis~\cite{Sun_2018,Wang_2018}. However, since human face can be considered as a more rigid object than the human body, the misalignment between the input and output images is limited and high quality images can be obtained without feature alignment.}
 
It is worth noticing that,
for discriminative tasks, other architectures have been proposed to deal with spatial deformations.
For instance, Jaderberg et al. \cite{jaderberg2015spatial}  propose a
 spatial transformer layer, which learns how to transform a feature map in a ``canonical'' view, conditioned on the feature map
itself.
However, only a global,  parametric  transformation can be learned (e.g., a global affine transformation), while in this paper we deal with  non-parametric deformations of articulated objects which cannot be described by means of a unique global affine transformation.

 Finally, our nearest-neighbour loss is similar to the
 perceptual loss  \cite{DBLP:conf/eccv/JohnsonAF16} and to the 
  style-transfer spatial-analogy approach  \cite{DBLP:journals/tog/LiaoYYHK17}. However, 
the perceptual loss,  based on an element-by-element difference computed in the feature map of an external classifier \cite{DBLP:conf/eccv/JohnsonAF16}, does not take into account spatial misalignments. On the other hand,   
  the patch-based similarity, adopted in \cite{DBLP:journals/tog/LiaoYYHK17} to compute a dense feature correspondence, is very computationally expensive and it is not used as a loss.

%% file: model.tex
\section{Deformable GANs} 
\label{architectures}

\begin{figure*}[t]\centering
\includegraphics[width=0.99\linewidth]{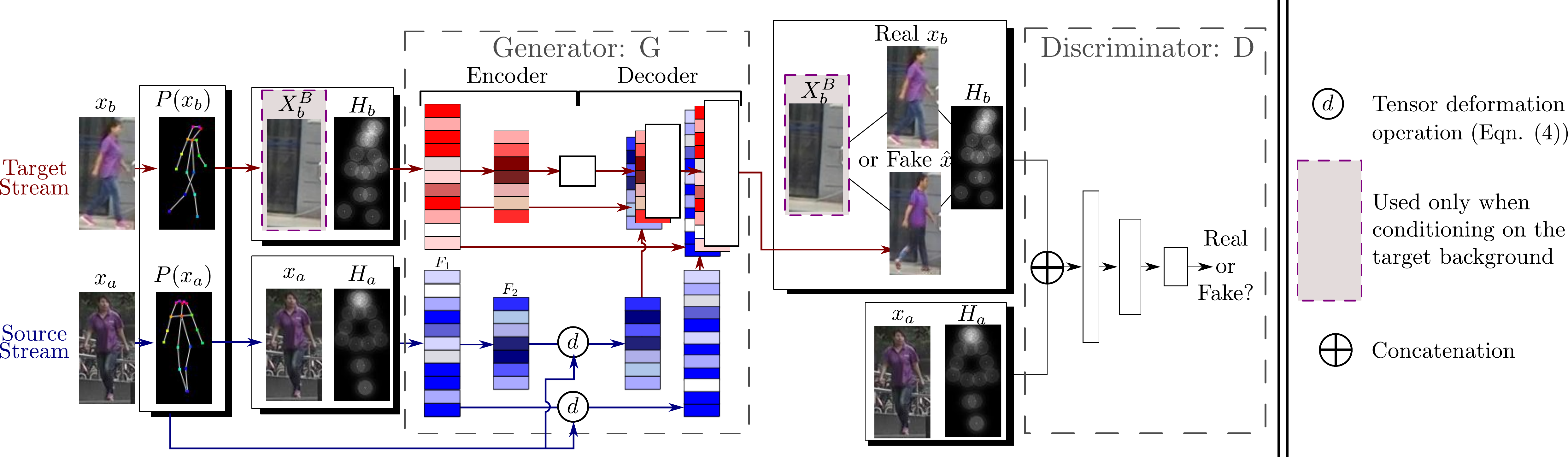}
\caption{A schematic representation of our network architectures. For the sake of clarity, in this figure we depict $P(\cdot)$ as a skeleton and each tensor $H$ as the average of its component matrices $H_j$ ($1 \leq j \leq k$). The white rectangles in the decoder represent the feature maps directly obtained using up-convolutional filters applied to the previous-layer maps. The reddish rectangles represent the feature maps ``shuttled'' by the skip connections in the target stream. Finally, blueish rectangles represent the deformed tensors ($d(F)$) ``shuttled'' by the deformable skip connections in the source stream.}
\vspace{-0.5cm}
\label{fig:pipeline}
\end{figure*}

At testing time our task, similarly to \cite{ma2017pose}, consists in generating an image $\hat{x}$ showing a person whose appearance (e.g., clothes, etc.) is similar to a conditioning appearance image $x_a$ but with a body pose similar to $P(x_b)$, where $x_b$ is a different image of the same person and $P(x) = (\mathbf{p}_1, ... \mathbf{p}_k)$ is a sequence of $k$ 2D points describing the locations of the human-body joints in $x$. 
In order to allow  a fair comparison with  \cite{ma2017pose} 
and other works, 
we use the same number  of joints ($k = 18$)
and we extract  $P()$ using the same Human Pose Estimator (HPE)  \cite{Cao} used in \cite{ma2017pose}. Note that this HPE is used both at testing and at training time, meaning that we do not use manually-annotated poses and the so extracted joint locations may have some localization errors or missing detections/false positives. At training time we use a dataset  ${\cal X} = \{ (x_a^{(i)}, x_b^{(i)}) \}_{i=1,...,N}$ containing pairs of conditioning-target images of the same person in different poses.
For each pair $(x_a, x_b)$, two poses $P(x_a)$ and $P(x_b)$ are extracted from the corresponding images and represented using two  tensors $H_a = H(P(x_a))$ and $H_b = H(P(x_b))$. Each tensor is composed of $k$ heat maps, where  $H_j$ ($1 \leq j \leq k$) is a 2D matrix of the same dimension as the original image. If $\mathbf{p}_j$ is the j-th joint location, then:

\begin{equation}
\label{eq.blurring}
H_j(\mathbf{p}) = exp\left(-\frac{\lVert \mathbf{p}- \mathbf{p}_j \rVert}{\sigma^2}\right),
\end{equation}

\noindent
with $\sigma = 6$ pixels (chosen with cross-validation).
Using
 blurring (Eq.~\eqref{eq.blurring}) instead of a binary map adopted in \cite{ma2017pose}, is useful to provide widespread information about the location $\mathbf{p}_j$.

The generator $G$ is fed with: (1) a noise vector $z$, 
 drawn from a noise distribution ${\cal Z}$ and implicitly provided using dropout \cite{pix2pix2016} 
and (2) the triplet $(x_a, H_a, H_b)$. Note that, at testing time, the target pose is known, thus $H(P(x_b))$ can be computed.
Note also that the joint locations in $x_a$ and  $H_a$ are spatially aligned (by construction), while in $H_b$ they are different.
Hence, differently from \cite{ma2017pose,pix2pix2016}, $H_b$ is not concatenated with the other input tensors. Indeed the convolutional-layer units in the encoder part of $G$ have a small receptive field which cannot capture large spatial displacements. For instance, when there is a large movement of a body limb in $x_b$ with respect to $x_a$, this limb is represented in different locations in $x_a$ and $H_b$ which may be too far apart  from each other to be captured by the receptive field of the convolutional  units. This is emphasized in the first layers of the encoder, which represent low-level information. Therefore, the convolutional filters cannot simultaneously process texture-level information (from $x_a$) and the corresponding pose information (from $H_b$). For this reason we independently process  $x_a$ and $H_a$ from $H_b$ in the encoder. 
Specifically, $x_a$ and $H_a$ are concatenated and processed
using the {\em source stream} of the encoder while  $H_b$ is processed by means of the {\em target stream},  without weight sharing  
(Fig.~\ref{fig:pipeline}).
The feature maps of the first stream are then fused with the layer-specific  feature maps of the second stream in the decoder  after a pose-driven spatial deformation performed by our deformable skip connections (see Sec.~\ref{skip-connections}).

Our discriminator network is based on the conditional, fully-convolutional discriminator proposed by
Isola et al. \cite{pix2pix2016}. In our case, $D$ takes as input 4 tensors: $(x_a, H_a, y, H_b)$, where either  $y =  x_b$ or $y =  \hat{x} = G(z, x_a, H_a, H_b)$ (see Fig.~\ref{fig:pipeline}).
These four tensors are concatenated  and then given as input to $D$. The  discriminator's output is a  scalar value indicating its confidence on the fact that  $y$ is a real image.

\subsection{Deformable skip connections}
\label{skip-connections}

As mentioned above  and similarly to  \cite{pix2pix2016},
the goal of the deformable skip connections is to ``shuttle'' local information from the encoder to the decoder part of $G$. The local information to be transferred is, generally speaking, contained in a tensor $F$, which represents the feature map activations of a given convolutional layer of the encoder.
However, differently from \cite{pix2pix2016}, we need to ``pick''
the information to shuttle taking into account  the object-shape deformation which is  described by the difference between $P(x_a)$ and $P(x_b)$.
To do so, we decompose the global deformation in a set of local affine transformations, defined using subsets of joints in $P(x_a)$ and $P(x_b)$.
Using these affine transformations together with local masks constructed using the specific joints, we deform the content of $F$ and then we use common skip connections
to copy the transformed tensor and concatenate it with the corresponding tensor in the destination layer (see  Fig.~\ref{fig:pipeline}).
Below we describe in more detail the whole pipeline.

\begin{figure}[t!]\centering
\includegraphics[angle=-90,width=0.9\linewidth]{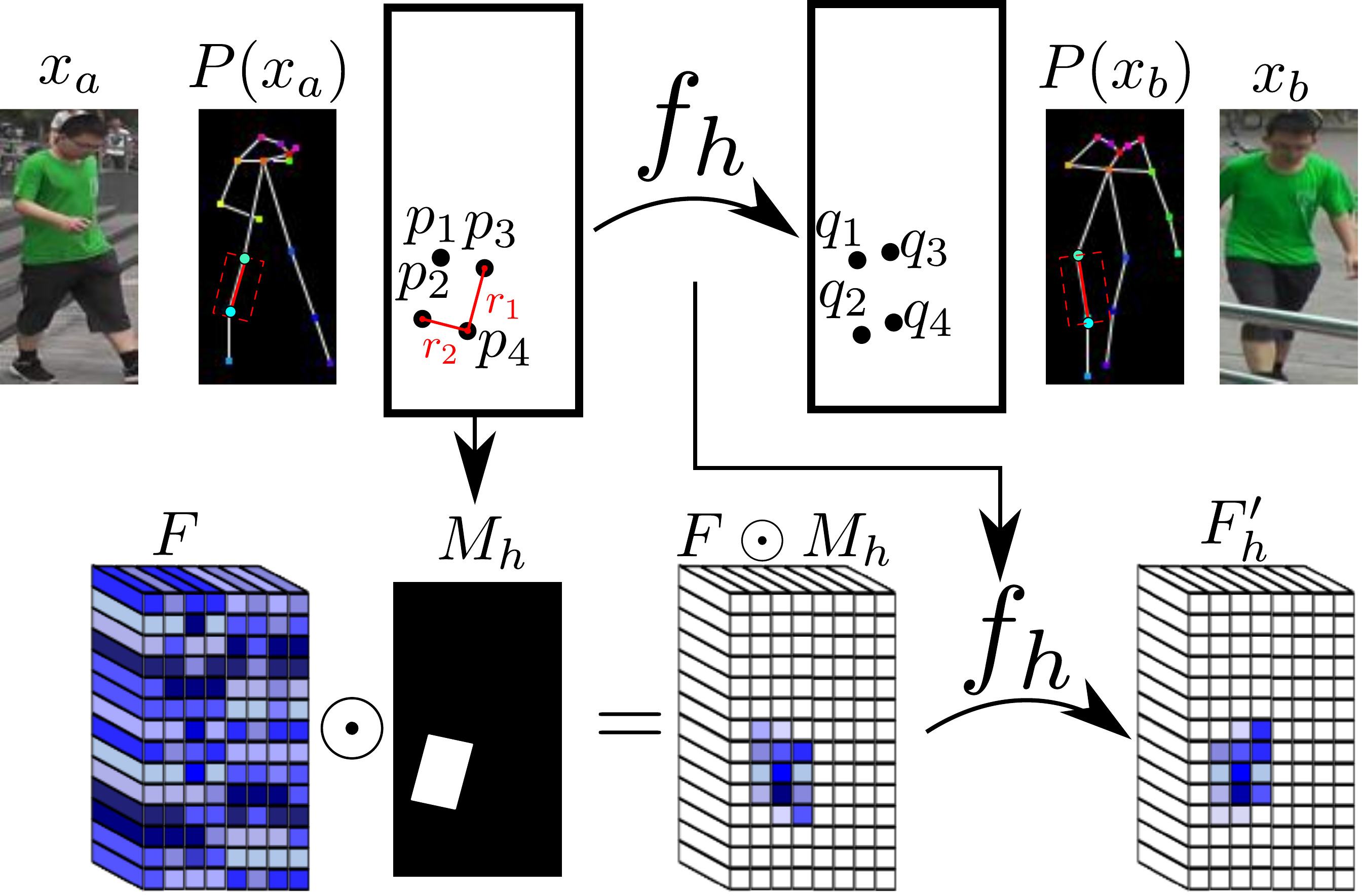}
\caption{For each specific body part, an  affine transformation $f_h$  is computed. This transformation is  used to ``move'' the feature-map content corresponding to that body part.}
\label{fig:affinepipeline}
\end{figure}

{\bf Decomposing an articulated body in a set of rigid sub-parts.} 
The human body is an articulated ``object'' which can be roughly decomposed into a set of rigid sub-parts. We chose 10 sub-parts: the head, the torso, the left/right upper/lower arm and the  left/right upper/lower leg. Each of them corresponds to a subset of the 18 joints defined by the HPE  \cite{Cao} we use for extracting $P()$. 
Using these  joint locations we can define rectangular regions which enclose the specific body part.
In case of  the head, the  region is simply chosen to be the axis-aligned enclosing rectangle of all the corresponding joints. For the torso, which is the largest area, we use a region which includes  the whole image, in such a way to shuttle  texture information for the background pixels.
Note that in Sec.~\ref{subSec:BGcontrol} we present an alternative way to generate background information.
Concerning the body limbs, each limb corresponds to only 2 joints. In this case we define a region to be a rotated rectangle whose major axis ($r_1$) corresponds to the line between these two joints, while the minor axis ($r_2$) is orthogonal to $r_1$ and with a length equal to one third of the mean of the torso's diagonals (this value is used for all the limbs). In Fig.~\ref{fig:affinepipeline} we show an example.
\addnote[Rhcom]{1}{
  Let $R_h^a = \{ \mathbf{p}_1, ..., \mathbf{p}_4 \}$ be the set of the 4 rectangle corners in $x_a$ defining the $h$-th body region ($1 \leq h \leq 10$).
 Note that these 4 corner points are not joint locations.
 Using $R_h^a$, we can compute a binary mask  $M_h(\mathbf{p})$ which is zero everywhere except the points $\mathbf{p}$ lying inside the rectangle area corresponding to $R_h^a$.}

 Moreover, let $R_h^b = \{ \mathbf{q}_1, ..., \mathbf{q}_4 \}$ be the corresponding rectangular region in $x_b$. Matching the points in $R_h^a$ with the corresponding points in $R_h^b$ we can compute the parameters of a body-part specific affine transformation (see below).
In  either  $x_a$ or $x_b$, some of the body regions can be occluded, truncated by the image borders or simply miss-detected by the HPE. In this case we leave  the corresponding region $R_h$ empty and the $h$-th affine transform is not computed (see below). 

Note  that our body-region definition  is the only human-specific part of the proposed approach. However, similar regions can be easily defined using the joints of other articulated objects such as those representing an animal body  or a human face.

{\bf Computing a set of affine transformations.} 
During the forward pass (i.e., both at training and at testing time) we decompose the global deformation of the conditioning pose with respect to the target pose by means of a set of local affine transformations, one per body region. Specifically, given $R_h^a$ in $x_a$ and $R_h^b$ in $x_b$ (see above), we compute the 6  parameters $\mathbf{k}_h$ of an affine transformation $f_h(\cdot ; \mathbf{k}_h)$
using Least Squares Error:

\begin{equation}
\min_{\mathbf{k}_h} \sum_{\mathbf{p}_j \in R_h^a, \mathbf{q}_j \in R_h^b} || \mathbf{q}_j - f_h(\mathbf{p}_j ; \mathbf{k}_h) ||^2_2
\end{equation}

The parameter vector 
$\mathbf{k}_h$ is computed using the original image resolution of $x_a$ and $x_b$ and then adapted to the specific resolution of each involved  feature map $F$.
Similarly, we compute scaled versions of each $M_h$.
In case either $R_h^a$ or $R_h^b$ is empty (i.e., when any of the specific body-region joints has not been detected using the HPE, see above), then we simply set   $M_h$ to be a matrix with all elements equal to 0 ($f_h$ is not computed).

 Note that  $(f_h(), M_h)$ and their lower-resolution variants need to be computed 
only once  per each pair of real images $(x_a, x_b) \in {\cal X}$ and, in case of the training phase, this can be done before starting training the networks (but in our current implementation this is done on the fly).

{\bf Combining affine transformations to approximate the  object deformation.} 
Once $(f_h(), M_h)$, $h= 1, ...,10$ are computed for the specific spatial resolution of a given tensor $F$, the latter can be transformed in order to approximate the global pose-dependent deformation.
Specifically, we first compute for each $h$:

\begin{equation}
  F'_h = f_h (F \odot M_h),
  \label{eq.F}
\end{equation}

\noindent
where $\odot$ is a point-wise multiplication
and $f_h(F(\mathbf{p}))$ is used to ``move'' all the channel values of $F$ corresponding to point 
$\mathbf{p}$.
\addnote[eq4c]{1}{Finally, we merge the resulting tensors treating each feature channel $c$ independently:
\begin{equation}
\label{eq.d-F}
d(F(\mathbf{p},c)) = max_{h = 1, ..., 10} F'_h(\mathbf{p},c).
\end{equation}
In other words, for each channel $c$ and each feature location $p$, we select the maximum value over the ten feature maps corresponding to the ten considered body-parts.~}
The rationale behind Eq.~\eqref{eq.d-F} is that, when two body regions partially overlap, the final deformed tensor $d(F)$ is obtained by picking the  maximum-activation values. \addnote[maxOrNot]{1}{We experimentally show the benefit of this hard-decision formulation over softer combinations such as averaging in Sec~\ref{Ablation}.}
\addnote[BGpreserve-eq]{1}{Note that, background is not modeled in Eqs.~\eqref{eq.F} and ~\eqref{eq.d-F} since there is no need for preserving it within the source stream. This point is further discussed in Sec.~\ref{subSec:BGcontrol}.}

\subsection{Training}
\label{Training}

$D$ and $G$ are trained using a combination of a standard conditional adversarial loss ${\cal L}_{cGAN}$ with our proposed nearest-neighbour loss ${\cal L}_{NN}$. Specifically, in our case 
${\cal L}_{cGAN}$ is given by:

\begin{equation}
\label{eq.GAN-loss}
\begin{array}{cc}
{\cal L}_{cGAN}(G,D)= &
\hspace{-7pt} \mathbb{E}_{(x_a,x_b) \in {\cal X}} [\log D(x_a, H_a, x_b, H_b)] + \\
 & \hspace{-40pt} \mathbb{E}_{(x_a,x_b) \in {\cal X}, z \in {\cal Z}} [\log ( 1 - D(x_a, H_a, \hat{x}, H_b ) )],
\end{array}
\end{equation}

\noindent
where $\hat{x} = G(z, x_a, H_a, H_b)$. 

Previous works on conditional GANs  combine the adversarial loss with either an $L_2$  \cite{DBLP:journals/corr/PathakKDDE16} or an $L_1$-based loss \cite{pix2pix2016,ma2017pose} which is used only for $G$. For instance, the $L_1$ distance computes a pixel-to-pixel difference between the generated and the real image, which, in our case, is:

\begin{equation}
L_1(\hat{x},x_b) =  ||\hat{x} - x_b ||_1.
\end{equation}

\noindent
However,
a well-known problem behind the use of $L_1$ and $L_2$ is the production of blurred images.
We hypothesize that this is also due to the inability of these losses to tolerate small spatial misalignments between $\hat{x}$ and $x_b$. For instance, suppose that 
$\hat{x}$, produced by $G$, is visually plausible and semantically similar to $x_b$, but the texture details on the clothes of the person in the two compared images are not pixel-to-pixel aligned. Both the $L_1$ and the $L_2$ losses will penalize this inexact pixel-level alignment, although not semantically important from the human point of view.
 Note that these misalignments do {\em not} depend on the global deformation between $x_a$ and $x_b$, because  $\hat{x}$ is supposed to have the same pose as $x_b$. In order to alleviate this problem, we propose to use a {\em nearest-neighbour} loss ${\cal L}_{NN}$ based on the following definition of image difference:

\begin{equation}
\label{eq.L-NN}
L_{NN} (\hat{x},x_b) = \sum_{\mathbf{p} \in \hat{x}}  min_{\mathbf{q} \in {\cal N}(\mathbf{p})} || g(\hat{x}(\mathbf{p})) -g(x_b (\mathbf{q})) ||_1,
\end{equation}

\noindent
where ${\cal N}(\mathbf{p})$ is a $n \times n$ local 
neighbourhood of point $\mathbf{p}$. 
$g(x (\mathbf{p}))$ is a vectorial representation of a patch around point $\mathbf{p}$ in image $x$, obtained using convolutional filters (see below for more details).
Note that $L_{NN}()$ is not a metric because it is not symmetric. 
In order to efficiently compute Eq.~\eqref{eq.L-NN}, we compare patches in $\hat{x}$ and $x_b$ using their representation
($g()$) in a  
convolutional map of an externally trained network.
In more detail, we use VGG-19 \cite{DBLP:journals/corr/SimonyanZ14a}, trained on ImageNet  and, specifically, its second convolutional layer (called $conv_{1\_2}$). The first two convolutional maps in VGG-19 ($conv_{1\_1}$ and  $conv_{1\_2}$) are both obtained using a convolutional stride equal to 1. For this reason, the feature map ($C_x$) of an image $x$ in  $conv_{1\_2}$ has the same resolution of the original image $x$. Exploiting this fact, we compute the nearest-neighbour field directly on  $conv_{1\_2}$, without losing spatial precision. 
Hence, we define: $g(x (\mathbf{p})) = C_x(\mathbf{p})$, which corresponds to the vector of all the channel values of $C_x$ with respect to the spatial position $\mathbf{p}$. $C_x(\mathbf{p})$ has a receptive field of $5 \times 5$ in $x$, thus effectively representing a patch of dimension $5 \times 5$ using a cascade of two convolutional layers interspersed by a non-linearity. Using $C_x$, Eq.~\eqref{eq.L-NN} becomes:

\begin{equation}
\label{eq.L-NN-conv}
L_{NN} (\hat{x},x_b) = \sum_{\mathbf{p} \in \hat{x}}  min_{\mathbf{q} \in {\cal N}( \mathbf{p})} ||C_{\hat{x}}(\mathbf{p}) - C_{x_b} (\mathbf{q}) ||_1.
\end{equation}

\noindent
 In Sec.~\ref{GPU},
 we show how \eqref{eq.L-NN-conv} can be efficiently implemented 
  using GPU-based parallel computing.
   The final $L_{NN}$-based loss is:

\begin{equation}
\label{eq.NN-loss}
{\cal L}_{NN}(G) = 
\mathbb{E}_{(x_a,x_b) \in {\cal X}, z \in {\cal Z}} L_{NN}(\hat{x},x_b).
\end{equation}

Combining Eq.~\eqref{eq.GAN-loss} and Eq.~\eqref{eq.NN-loss} we obtain our  objective:

\begin{equation}
\label{eq.objective}
G^* = \arg \min_G \max_D {\cal L}_{cGAN}(G,D) + \lambda {\cal L}_{NN}(G),
\end{equation}

\noindent
with $\lambda = 0.01$ used
 in all our experiments. The value of $\lambda$ is small because it acts as a normalization factor in  Eq.~\eqref{eq.L-NN-conv} with respect to the number of channels in $C_x$ and the number of pixels in  $\hat{x}$ (more details in  
 Sec.~\ref{GPU}).

\subsection{Conditioning on the Background}
\label{subSec:BGcontrol}

We now introduce a third (optional) conditioning variable: the background. Controlling the background generation has two main practical interests. First, in the context of data augmentation, we can increase the diversity of the generated data by using different background images.
 Second, when we aim at generating image sequences (e.g., short videos), a temporally coherent background is helpful. Consequently, in this section we show how to extend the proposed method in order to generate the background area conditioned on a given, third input image.

Formally speaking, the output image $\hat{x}$ should be conditioned (also) on a target background image 
$x^B_b$. In the generator, this is simply obtained by concatenating  $H_b$ with 
$x^B_b$ (see Fig.~\ref{fig:pipeline}). In this way,  background information can be provided to the decoder using the target stream.
 \addnote[BGpreserve]{1}{Note that, since the background image is part of the target stream (in red in Fig.~\ref{fig:pipeline}), the deformable skip connections defined in Eq. \eqref{eq.F} are not applied to the image features extracted from the background image. Therefore, there is no need to specifically modify Eq.~\eqref{eq.F} to handle background conditioning.} Similarly, the discriminator network takes $x^B_b$ as an additional input, which is concatenated with the other input images. 
 Therefore, the discriminator can detect whether the background of $\hat{x}$  corresponds to the conditioning background image $x^B_b$ and force the generator to output images with the desired background.

Training is performed using video sequences from which a background image can be easily extracted. More in details, 
we used  the PRW dataset \cite{zheng2016person} 
which contains a set of videos annotated with the bounding box of each tracked person.
When we train our networks with background conditioning information, we extract $x_a$ and $x_b$ from a person track in two different frames which may come from different
videos. 
Then,  $x^B_b$ is obtained 
by choosing at random a frame in the same video of $x_b$
 with no bounding box overlapping with the area corresponding to $x_b$.
Note that in the PRW dataset
the cameras are static and 
 the background objects do not move during the video (except for a few objects as, for instance, bikes).

%% file: implementation.tex
\section{Implementation details}
\label{sec:impDetails}

In this section we provide additional technical details of our proposed method. We first show how the proposed nearest-neighbour loss can be efficiently computed exploiting optimized matrix-multiplications typically used in GPU-based programming. Second, we show how to use the symmetry of the human body in order to handle possible missing/non-detected body parts. Finally, we report the details of the architectures and the training procedure used in our experiments.
\subsection{Nearest-neighbour loss implementation}
\label{GPU}

Our proposed nearest-neighbour loss 
is based on $L_{NN} (\hat{x}, x_b)$ given in Eq.~\eqref{eq.L-NN-conv}.
In that equation, for each point $\mathbf{p}$ in $\hat{x}$, the ``most similar'' (in the $C_x$-based feature space) point $\mathbf{q}$ in $x_b$ needs to be searched for in a $n \times n$ neighborhood of $\mathbf{p}$.
This operation may be time consuming if implemented using sequential computing (i.e., using a ``\texttt{for loop}'').
We show how this computation can be 
sped-up by exploiting GPU-based parallel computing in which different tensors are processed simultaneously. 

Given  $C_{x_b}$, we  compute $n^2$ 
shifted versions of $C_{x_b}$: $\{C_{x_b}^{(i, j)}\}$, where $(i,j)$ is a translation offset ranging in a relative $n \times n$ neighborhood
($i,j \in \{ - \frac{n-1}{2}, ..., + \frac{n-1}{2} \}$)  
and $C_{x_b}^{(i, j)}$
is filled with $+\infty$ in the borders. Using this translated versions of  $C_{x_b}$,
we compute $n^2$ corresponding difference tensors $\{D^{(i, j)}\}$, where:

\begin{equation}
D^{(i, j)} = |C_{\hat{x}} - C_{x_b}^{(i, j)}| 
\end{equation}

\noindent
and the difference is computed element-wise. $D^{(i, j)} (\mathbf{p})$ contains the channel-by-channel  absolute difference between 
 $C_{\hat{x}} (\mathbf{p})$ and $C_{x_b}(\mathbf{p} + (i, j))$.
 Then, for each $D^{(i, j)}$, we sum all the channel-based differences obtaining:  

\begin{equation} 
\label{eq.S}
 S^{(i, j)} = \sum_c D^{(i, j)}(c),
\end{equation} 

\noindent 
 where $c$ ranges over all the channels and the sum is performed  
pointwise. 
$S^{(i, j)}$ is a matrix of scalar values, each value representing the $L_1$ norm of the difference between a point $\mathbf{p}$ in $C_{\hat{x}}$ and a corresponding point $\mathbf{p} + (i, j)$ in  $C_{x_b}$:

\begin{equation}
S^{(i, j)}(\mathbf{p}) = ||C_{\hat{x}} (\mathbf{p}) - C_{x_b}(\mathbf{p} + (i, j)) ||_1.
\end{equation}

For each point $\mathbf{p}$, we can now compute its best match in a local neighbourhood of  $C_{x_b}$ simply using:

\begin{equation}
\label{eq.M}
M(\mathbf{p}) =   min_{(i, j)} S^{(i, j)}(\mathbf{p}).
\end{equation}

Finally,
Eq.~\eqref{eq.L-NN-conv} becomes:

\begin{equation}
\label{eq.M-sum}
L_{NN}(\hat{x},x_b) =   \sum_{\mathbf{p}} M(\mathbf{p}).
\end{equation}

Since we do not normalize  Eq.~\eqref{eq.S} by the number of  channels nor   Eq.~\eqref{eq.M-sum} by the number of pixels,
the final value $L_{NN}(\hat{x},x_b)$  is usually very high. For this reason we use a small value $\lambda = 0.01$ in Eq.~\eqref{eq.objective} when weighting ${\cal L}_{NN}$ with respect to ${\cal L}_{cGAN}$.

\subsection{Exploiting the human-body symmetry}
\label{symmetry}

As mentioned in Sec.~\ref{skip-connections}, we decompose the human body in 10 rigid sub-parts: the head, the torso and 8 limbs (left/right upper/lower arm, etc.). When one of the joints corresponding to one of these body-parts has not been detected by the HPE, the corresponding region and affine transformation are not computed and  the region-mask is filled with 0. This can happen because of either that region is not visible in the input image or because of false-detections of the HPE.  However, when the missing region involves a limb (e.g., the right-upper arm) whose symmetric body part has been detected (e.g., the left-upper arm), we can ``copy'' information from the ``twin'' part.
In more detail, suppose for instance that the region corresponding to the right-upper arm in the conditioning appearance image is $R_{rua}^a$ and this region is empty because of one of the above reasons. Moreover, suppose that $R_{rua}^b$ is the corresponding (non-empty) region in $x_b$ and that $R_{lua}^a$
is the (non-empty)  left-upper arm region in $x_a$. We simply set: $R_{rua}^a := R_{lua}^a$ and we compute $f_{rua}$ as usual, using the (now, no more empty) region $R_{rua}^a$ together with  $R_{rua}^b$.

\subsection{Network and Training details}
\label{sec:Network-details}

We train $G$ and $D$ for 90k iterations, with the Adam optimizer (learning rate: $2 * 10^{-4}$, $\beta_1 = 0.5$, $\beta_2 =0.999$). Following \cite{pix2pix2016} we use instance normalization \cite{DBLP:journals/corr/UlyanovVL16}. 
In the following we denote with: 
(1) $C_m^s$  a convolution-ReLU layer with $m$ filters and stride $s$,
(2)  $CN_m^s$  the same as $C_m^s$ with instance normalization before ReLU and
(3) $CD_m^s$ the same as $CN_m^s$  with the addition of  dropout at rate $50\%$.
Differently from \cite{pix2pix2016}, we use dropout only at training time.
The encoder part of the generator is given by two streams (Fig.~\ref{fig:pipeline}), each of which is composed of the following sequence of layers:

$CN_{64}^1 - CN_{128}^2 - CN_{256}^2 - CN_{512}^2 - CN_{512}^2 - CN_{512}^2$.

\noindent
The decoder part of the generator is given by:

$CD_{512}^2 - CD_{512}^2 - CD_{512}^2 - CN_{256}^2 - CN_{128}^2 - C_{3}^1$.

\noindent
In the last  layer,  ReLU is replaced with $tanh$.

The discriminator architecture is:

$C_{64}^2 - C_{128}^2 - C_{256}^2 - C_{512}^2 - C_{1}^2$,

\noindent
where the  ReLU of the last  layer  is replaced with $sigmoid$.

The generator for the DeepFashion dataset has one additional convolution block ($CN_{512}^2$)
both in the encoder and in the decoder, because images in this dataset  have a  higher resolution.

%% file: experimentation.tex
\section{Experiments}
\label{Experiments}

In this section we compare our method with other state-of-the-art person generation approaches, both qualitatively and quantitatively and we show an ablation study. Since quantitative evaluation of generative methods is still an open research problem, we adopt different criteria, which can be summarized in: (1) the evaluation protocols suggested by Ma et al. \cite{ma2017pose}, (2) human judgements and (3) experiments based on Re-ID training with data-augmentation.
Note that in all but the qualitative experiments shown in Sec.~\ref{exp:BGcontrol} we do {\em not} use the background conditioning information. Indeed, since most of the methods we compare with do not use additional background conditioning information, we also avoided this for a fair comparison.

\subsection{Datasets}
\label{expe:datasets}

The person Re-ID  Market-1501 dataset \cite{zheng2015scalable} contains 32,668 images of 1,501 persons captured from 6 different surveillance cameras. This dataset is challenging because of the low-resolution images (128$\times$64) and the high diversity in pose, illumination, background and viewpoint. To train our model, we need pairs of images of the same person in two different poses. As this dataset is relatively noisy, we first automatically remove those images in which no human body is detected using the HPE, leading to 263,631 training pairs.
For testing, following  \cite{ma2017pose}, we  randomly select 12,000 pairs. No person is in common between the training and the test split.
 
The DeepFashion dataset ({\em In-shop Clothes Retrieval Benchmark}) \cite{liu2016deepfashion} is composed of 52,712 clothes images, matched each other in order to form  200,000 pairs of identical clothes with two different poses and/or scales of the persons wearing these clothes. 
The images have a resolution of 256$\times$256 pixels. 
Following the training/test split adopted in \cite{ma2017pose}, we create pairs of images, each pair depicting the same person with identical clothes  but in different poses.
After removing those images in which the HPE does not detect any human body, we finally 
 collect 89,262 pairs for training and 12,000 pairs for testing.

\subsection{Metrics}
\label{Metrics}
Evaluation  in the context of generation tasks is a problem in itself. In our experiments 
 we adopt a redundancy of  metrics and,
following \cite{ma2017pose}, we use:  Structural Similarity (\emph{SSIM}) \cite{wang2004image},  Inception Score (\emph{IS}) \cite{salimans2016improved} and their corresponding masked versions \emph{mask-SSIM} and \emph{mask-IS} \cite{ma2017pose}. The latter are obtained by masking-out the image background and the rationale behind this is that, since no background information of the target image is input to $G$, the network cannot guess what the target background looks like
(remember that we do {\em not} use background conditioning in these experiments, see above).  
Note that the evaluation masks we use to compute both the mask-IS and the mask-SSIM values do not correspond to the masks ($\{M_h\}$) we use for training. The evaluation masks have been built following the procedure proposed in \cite{ma2017pose} and adopted in that work for both  training and evaluation. Consequently, the mask-based metrics may be biased in favour of their method.
Moreover, we observe that
 the IS metrics  \cite{salimans2016improved}, based on the entropy computed over the classification neurons of an external classifier 
\cite{DBLP:journals/corr/SzegedyVISW15}, 
is not very suitable for
domains with only one object class (the person class in this case). 
 For this reason we propose to  use an additional metric that we call Detection Score (\emph{DS}). Similarly to the classification-based metrics {\em FCN-score}, used in \cite{pix2pix2016}, DS is based on the detection outcome of
the state-of-the-art object detector SSD \cite{liu2016ssd},
trained on Pascal VOC  \cite{pascal-voc-2007} (and not fine-tuned on our datasets).
At testing time, we use the person-class detection scores of SSD 
computed on each generated image $\hat{x}$.
 $DS(\hat{x})$ corresponds to the maximum-score box of SSD on $\hat{x}$ and the final DS value is computed by averaging the scores of all the generated images. In other words, DS measures the confidence of a person detector on the presence of a person in the image. Given the high accuracy of SSD in the challenging Pascal VOC  dataset \cite{liu2016ssd}, we believe that it can be used as a good measure of how much realistic (person-like) is a generated image.

Finally, in our tables we also include  the value of each metric computed using the {\em real} images of the test set. Since these  values are computed on real data, they can be considered as a sort of an upper-bound to the results a generator can obtain. However, these  values are not actual upper bounds in the strict sense: for instance the DS metric on the real datasets is not 1 because of SSD failures.

\begin{table*}[h!]
\caption{Comparison with the state of the art. 
}
\centering
\begin{tabular}{l|ccccc|ccc}
  \toprule
  &\multicolumn{5}{c|}{Market-1501}&\multicolumn{3}{c}{DeepFashion}\\
 Model &\emph{SSIM} & \emph{IS}&\emph{mask-SSIM} & \emph{mask-IS} &\emph{DS}& \emph{SSIM} & \emph{IS} &\emph{DS}\\
\midrule
Ma et al. \cite{ma2017pose} &\bf$0.253$ & $3.460$ & $0.792$ & $3.435$& $0.39$ &$0.762$ & $3.090$ & $0.95$  \\
Ma et al. \cite{ma2018disentangled} &\bf$0.099$ & $\bf3.483$ & $0.614$ & $3.491$& $0.61$ &$0.614$ & $3.228$ & -  \\
Esser et al. \cite{esser2018variational} &$\bf0.353$ & $3.214$ & $0.787$ & $3.249$& $\bf 0.83$  &$\bf0.786$ & $3.087$ & $0.94$  \\
\emph{Ours}&$0.290$ & $3.185$ & $\bf0.805$ & $3.502$& $0.72$ &$0.756$ & $\bf 3.439$ & $\bf 0.96$ \\
\midrule
\emph{Real-Data}&$1.00$ & $3.86$ & $1.00$ & \comRev{$\bf3.76$}& $0.74$ &$1.000$ & $3.898$ & $0.98$ \\
\bottomrule
\end{tabular}
\label{tab:result}
\end{table*}

\subsection{Comparison with previous work}
\label{Comparison}

In this section we qualitatively and quantitatively compare our method with state-of-the-art person generation approaches.

\begin{figure}[h]
  \centering
  \setlength\tabcolsep{1.0pt}
\begin{tabular}{ccccc}
  $x_a$ & $x_b$ & \small\emph{Full (ours)}& Esser et al. \cite{esser2018variational} &Ma et al. \cite{ma2017pose}\\

\includegraphics[width=0.11\columnwidth]{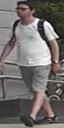}
&\includegraphics[width=0.11\columnwidth]{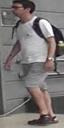}
&\includegraphics[width=0.11\columnwidth]{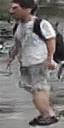}
&\includegraphics[width=0.11\columnwidth]{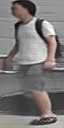}
&\includegraphics[width=0.11\columnwidth]{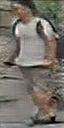}
\\
\includegraphics[width=0.11\columnwidth]{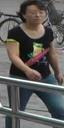}
&\includegraphics[width=0.11\columnwidth]{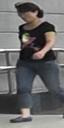}
&\includegraphics[width=0.11\columnwidth]{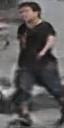}
&\includegraphics[width=0.11\columnwidth]{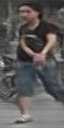}
&\includegraphics[width=0.11\columnwidth]{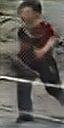}
\\
\includegraphics[width=0.11\columnwidth]{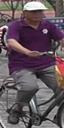}
&\includegraphics[width=0.11\columnwidth]{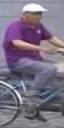}
&\includegraphics[width=0.11\columnwidth]{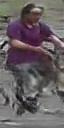}
&\includegraphics[width=0.11\columnwidth]{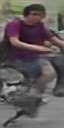}
&\includegraphics[width=0.11\columnwidth]{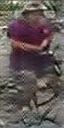}
\\
\includegraphics[width=0.11\columnwidth]{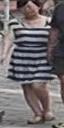}
&\includegraphics[width=0.11\columnwidth]{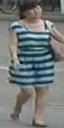}
&\includegraphics[width=0.11\columnwidth]{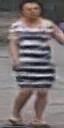}
&\includegraphics[width=0.11\columnwidth]{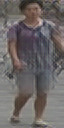}
&\includegraphics[width=0.11\columnwidth]{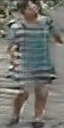}
\\
\includegraphics[width=0.11\columnwidth]{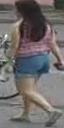}
&\includegraphics[width=0.11\columnwidth]{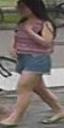}
&\includegraphics[width=0.11\columnwidth]{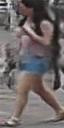}
&\includegraphics[width=0.11\columnwidth]{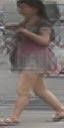}
&\includegraphics[width=0.11\columnwidth]{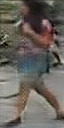}

\end{tabular}
  \caption{A qualitative comparison on the Market-1501 dataset between our approach and \cite{esser2018variational} and  \cite{ma2017pose}. Columns 1 and 2  show the (testing) conditioning appearance and pose image, respectively, which are used as reference by all methods. Columns 3, 4 and 5 respectively show the images generated by our full-pipeline and by \cite{esser2018variational} and  \cite{ma2017pose}.}
\label{fig:comparison-Market-2}
\end{figure}

{\bf Qualitative comparison.}
Fig.~\ref{fig:comparison-Market-2} shows the results on the Market-1501 dataset. 
Comparing the images generated by our full-pipeline with the corresponding images generated by the full-pipeline presented in \cite{ma2017pose}, most of the times our results are more realistic, sharper and with local details (e.g., the clothes  texture or the face characteristics) more similar to the details of the conditioning appearance image. 
In all the examples of Fig.~\ref{fig:comparison-Market-2} the method proposed in \cite{ma2017pose} produced images either containing more artefacts or more blurred than our corresponding images. Concerning the approach proposed in \cite{esser2018variational}, we observe that the generated images are 
sometimes more realistic than ours (e.g., rows 1 and 3).
However, the approach proposed in \cite{esser2018variational} is less effective in preserving the specific details of the conditioning appearance image. For instance, in the last row,  our method better preserves  the blue color of the shorts. Similarly, in the fourth row, the strips of the dress are well generated by our approach but not by \cite{esser2018variational}.

\begin{figure}[h]
  \centering
  \setlength\tabcolsep{1.0pt}
\begin{tabular}{ccccc}
$x_a$ & $x_b$ & \small\emph{Full (ours)}& Esser et al. \cite{esser2018variational} &Ma et al. \cite{ma2017pose}\\ 
\includegraphics[width=0.12\columnwidth]{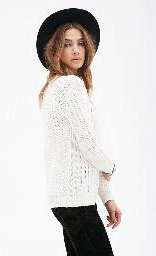}
&\includegraphics[width=0.12\columnwidth]{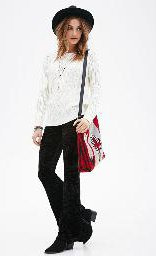}
&\includegraphics[width=0.12\columnwidth]{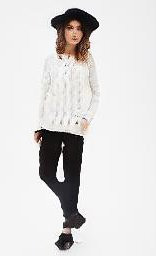}
&\includegraphics[width=0.12\columnwidth]{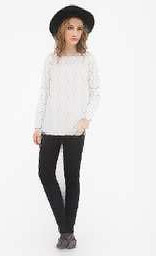}
&\includegraphics[width=0.12\columnwidth]{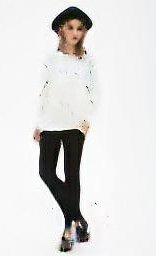}

\\
\includegraphics[width=0.12\columnwidth]{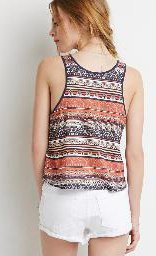}
&\includegraphics[width=0.12\columnwidth]{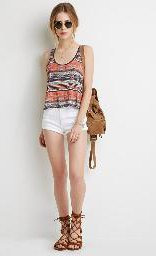}
&\includegraphics[width=0.12\columnwidth]{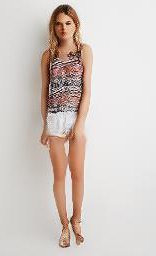}
&\includegraphics[width=0.12\columnwidth]{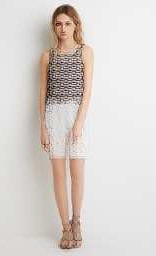}
&\includegraphics[width=0.12\columnwidth]{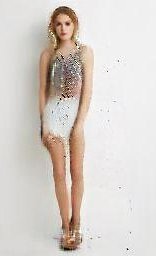}
\\
\includegraphics[width=0.12\columnwidth]{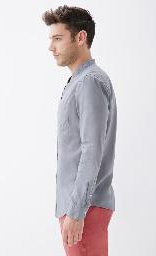}
&\includegraphics[width=0.12\columnwidth]{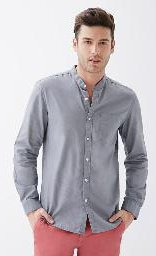}
&\includegraphics[width=0.12\columnwidth]{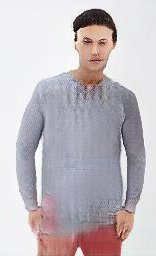}
&\includegraphics[width=0.12\columnwidth]{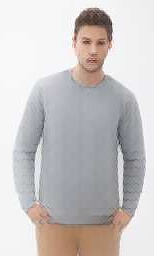}
&\includegraphics[width=0.12\columnwidth]{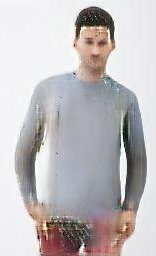}
\\
\includegraphics[width=0.12\columnwidth]{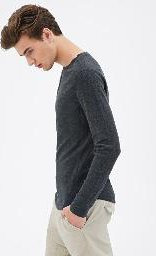}
&\includegraphics[width=0.12\columnwidth]{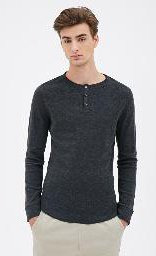}
&\includegraphics[width=0.12\columnwidth]{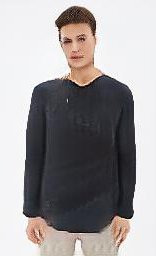}
&\includegraphics[width=0.12\columnwidth]{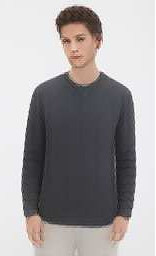}
&\includegraphics[width=0.12\columnwidth]{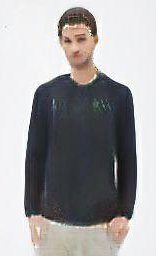}
\end{tabular}
  \caption{A qualitative comparison on the DeepFashion dataset between our approach and the results obtained by \cite{esser2018variational} and  \cite{ma2017pose}.}
\label{fig:comparison-Fashion-2}
\end{figure}

Fig.~\ref{fig:comparison-Fashion-2} shows the results on the DeepFashion dataset. Also in this case, comparing our results with \cite{ma2017pose}, most of the times ours  look more realistic or closer to the details of the conditioning appearance image.  For instance, the third row of Fig.~\ref{fig:comparison-Fashion-2} shows many artefacts in the image generated by \cite{ma2017pose}. Additionally, we see in the first two rows that our method more effectively transfers the details of the corresponding clothes textures.
Concerning  \cite{esser2018variational}, and similarly to the Market-1501 dataset,  the generated images   are usually smoother and sometimes more realistic that ours. However, the appearance details look more generic and less conditioned on the specific details contained in $x_a$.
For instance, the shape of the shorts in the second row or the color  of the trousers in the third row are less similar to the details of the corresponding appearance image $x_a$ with respect to our results.

We believe that this qualitative comparison 
 shows that the combination of the proposed deformable skip-connections and the nearest-neighbour loss produced the desired effect to ``capture'' and transfer the correct local details from the conditioning appearance image to the generated image.
 Transferring local information while simultaneously taking into account the global pose deformation is a difficult task which can more hardly be implemented using ``standard''
 U-Net based generators as those adopted in \cite{ma2017pose,esser2018variational}.
We also believe that this comparison shows that our method is better able than \cite{esser2018variational} to transfer person-specific details. This observation based on a qualitative comparison is confirmed by the quantitative experiments using different Re-ID systems in Sec.~\ref{PersonRe-ID}. A significant, drastic {\em decrease} of the accuracy of all the tested Re-ID systems, obtained when using \cite{esser2018variational} for data augmentation, indirectly shows that the generated images are too generic to be used to populate a Re-ID training set.

{\bf Quantitative comparison.}
Using the  metrics presented in 
Sec.~\ref{Metrics} we perform the
 quantitative evaluation shown in Tab.~\ref{tab:result}. 
Since background in this dataset is uniform and trivial to be reproduced, the mask-based metrics are not reported in the papers of the competitor methods for the DeepFashion dataset.
Concerning the DS metrics,
we used the publicly available code and  network weights released by the authors of 
\cite{ma2017pose,ma2018disentangled,esser2018variational} 
in order to generate new images according to the common testing protocol  and  ran the SSD detector to get the DS values. 
Note that the DS metric is not reported for \cite{ma2018disentangled} because the authors have not released the code nor the generated images for the DeepFashion dataset.
On the Market-1501 dataset our method reports the highest performance according to the mask-SSIM and the mask-IS  metrics. 
Note that, except \cite{ma2018disentangled}, none of the methods, including ours, is explicitly conditioned on background information, thus the mask-based metrics purely compare the region under conditioning.
Ranking the methods according to  the non-masked metrics is less easy.
Specifically, our  DS values are much higher than those obtained by \cite{ma2017pose}
but lower than the scores obtained by \cite{esser2018variational}. A bit surprising, the DS scores obtained using \cite{esser2018variational} are even higher than the values obtained using  real data. We presume this is due to the fact that the images generated by \cite{esser2018variational} look very realistic but probably they are relatively easy for a detector to be recognized, lacking sufficient inter-person variability. The experiments performed in Sec.~\ref{PersonRe-ID} indirectly confirm this interpretation.
Conversely, on the DeepFashion dataset, our approach ranks the first with respect to the IS and the 
DS metrics 
and the third with respect to the SSIM metrics. 
 This incoherence in the rankings 
 illustrates that no final conclusion can be drawn using only the metrics presented in 
Sec.~\ref{Metrics}. For this reason, we extend the comparison performing \comRev{two user studies}
(Sec.~\ref{Userstudy}) and experiments based on person Re-ID (Sec.~\ref{PersonRe-ID}).

\subsection{User study}
\label{Userstudy}

\addnote[user1]{1}{In order to further compare our approach with state of  the art methods, 
 we implement two  different user studies. On the one hand, we 
follow the protocol of Ma et al. \cite{ma2017pose}.}~ For each dataset, we show
55 real  and 55 generated images in a random order to 30 users for one second.
Differently from Ma et al. \cite{ma2017pose}, who used Amazon Mechanical Turk (AMT),
we used 
 ``expert'' (voluntary) users: PhD students and Post-docs working in Computer Vision and belonging to two different departments.
 We believe that expert users, who are familiar with GAN-like images,
 can more easily 
 distinguish  real  from fake images,  thus confusing our users is potentially a more difficult task for our GAN.
In Tab.~\ref{tab:user-study} we show our results\footnote{$R2G$ means $\#$Real images rated as generated / $\#$Real images; 
$G2R$ means $\#$Generated images rated as Real / $\#$Generated images.} 
together with the results reported in \cite{ma2017pose}. We believe these results can be compared to each other because were obtained using the same experimental protocol, although
they have been obtained using different sets of users. No user-study is reported in \cite{esser2018variational,ma2018disentangled}.
Tab.~\ref{tab:user-study} 
confirms the significant quality boost of our images with respect to the images produced in
\cite{ma2017pose}. For instance, on the  Market-1501 dataset, the $G2R$ human ``confusion'' is one order of magnitude higher than in \cite{ma2017pose}.

\begin{table}[h!]
\caption{User study ($\%$).
$(*)$ These results are reported in \cite{ma2017pose} and refer to a similar study with AMT users. } 
\centering
\begin{tabular}{lcccc}
  \toprule
  &\multicolumn{2}{c}{Market-1501}&\multicolumn{2}{c}{DeepFashion}\\
\midrule
  Model &\emph{R2G} & \emph{G2R} & \emph{R2G} & \emph{G2R} \\
\midrule
Ma et al. \cite{ma2017pose}   $(*)$        & 11.2       & 5.5        & 9.2       & 14.9 \\
\emph{Ours}             & \bf 22.67     & \bf 50.24  &  \bf 12.42   & \bf 24.61 \\
\bottomrule
\end{tabular}
\label{tab:user-study}
\end{table}

\addnote[user2]{1}{On the other hand, we propose to directly compare the images generated by our method and by the methods of \cite{ma2017pose,esser2018variational}. Specifically, we randomly chose a source image and a target pose. We then show to the user the source image and three generated images (one per method). The user is asked to select the most realistic image of the person in the source image. By displaying the source image, we aim at evaluating both realism and appearance transfer. Using AMT, we ask 10 users to reapeat this evaluation on 50 different source images for each dataset. Results are reported in Tab.\ref{tab:user-study-direct}. We observe that for both datasets, our method is chosen most frequently (in about $45\%$ of the cases). When comparing with the performances of \cite{ma2017pose}, our approach reaches a preference percentage that is about twice higher. This result is well in line with the first user study reported in Tab.\ref{tab:user-study}. Overall, it shows again that our approach outperforms other methods on both datasets.}


\begin{table}[h!]
\caption{\comRev{User study based on direct comparisons: we report user preference in $\%$.} }
\centering
\begin{tabular}{lcc}
  \toprule
  Model&Market-1501&DeepFashion\\
\midrule
Ma et al. \cite{ma2017pose}  & 23.8& 19.4\\
Esser et al. \cite{esser2018variational}             & 30.0& 35.8\\
\emph{Ours}        &     \bf46.2& \bf 44.8\\
\bottomrule
\end{tabular}
\label{tab:user-study-direct}
\end{table}

\input{exp-RID}

\subsection{Qualitative evaluation of the background-based conditioning}

We provide in this section a qualitative evaluation of the background conditioning variant of our method 
presented in Sec.~\ref{subSec:BGcontrol}. As aforementioned, for a fair comparison, we have {\em not} used background conditioning information in the experiments in Secs.~\ref{Comparison},~\ref{Userstudy} and \ref{PersonRe-ID}, since all the other methods we compare with do not use additional background  information
in their generation process.

\begin{figure}[h]
  \centering
  \setlength\tabcolsep{0.5pt}
\begin{tabular}{p{0.105\columnwidth}p{0.105\columnwidth}ccccc||cc}

  \vspace{-1.2cm}\centering$x_a$&\vspace{-1.7cm} $~~~x_b^B$\newline \centering{\Large$\diagdown$} \newline $P(x_b)~~$&
  \includegraphics[width=0.105\columnwidth]{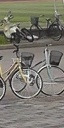}&
\includegraphics[width=0.105\columnwidth]{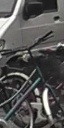}&
\includegraphics[width=0.105\columnwidth]{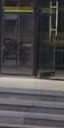}&
\includegraphics[width=0.105\columnwidth]{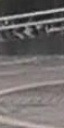}&
\includegraphics[width=0.105\columnwidth]{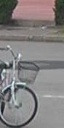}&
\includegraphics[width=0.105\columnwidth]{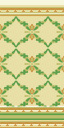}&
\includegraphics[width=0.105\columnwidth]{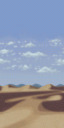}
\\
\includegraphics[width=0.105\columnwidth]{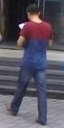}&
\includegraphics[width=0.105\columnwidth]{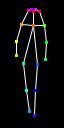}&
\includegraphics[width=0.105\columnwidth]{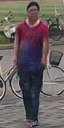}&
\includegraphics[width=0.105\columnwidth]{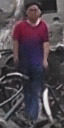}&
\includegraphics[width=0.105\columnwidth]{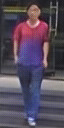}&
\includegraphics[width=0.105\columnwidth]{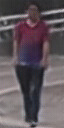}&
\includegraphics[width=0.105\columnwidth]{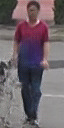}&
\includegraphics[width=0.105\columnwidth]{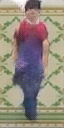}&
\includegraphics[width=0.105\columnwidth]{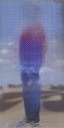}
\\
\includegraphics[width=0.105\columnwidth]{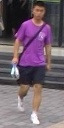}&
\includegraphics[width=0.105\columnwidth]{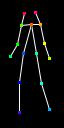}&
\includegraphics[width=0.105\columnwidth]{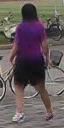}&
\includegraphics[width=0.105\columnwidth]{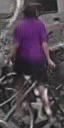}&
\includegraphics[width=0.105\columnwidth]{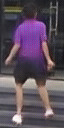}&
\includegraphics[width=0.105\columnwidth]{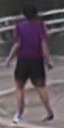}&
\includegraphics[width=0.105\columnwidth]{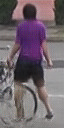}&
\includegraphics[width=0.105\columnwidth]{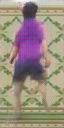}&
\includegraphics[width=0.105\columnwidth]{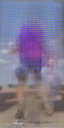}
\\

\includegraphics[width=0.105\columnwidth]{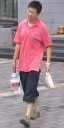}&
\includegraphics[width=0.105\columnwidth]{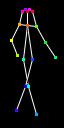}&
\includegraphics[width=0.105\columnwidth]{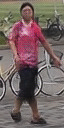}&
\includegraphics[width=0.105\columnwidth]{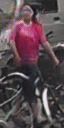}&
\includegraphics[width=0.105\columnwidth]{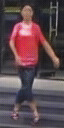}&
\includegraphics[width=0.105\columnwidth]{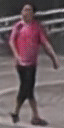}&
\includegraphics[width=0.105\columnwidth]{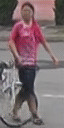}&
\includegraphics[width=0.105\columnwidth]{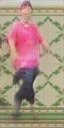}&
\includegraphics[width=0.105\columnwidth]{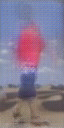}\\

\end{tabular}
  \caption{Qualitative results on the PRW dataset when conditioning on the background. We use three different pairs of conditioning appearance image $x_a$ and target pose $P(x_b)$. \comRev{For each pair, we use five different target background images $x^B_b$ extracted from the PRW dataset and 2 background images that are visually really different from the PRW dataset backgrounds (last two columns).}}
\label{fig:prwSequence}
\end{figure}

In Fig.~\ref{fig:prwSequence}, we show some qualitative results combining different triplets of conditioning variables $(x_a, P(x_b), x^B_b)$. \addnote[expefailure]{1}{For each input pair, we employ seven different target background images $x^B_b$. The first five background images are extracted from the PRW dataset and the last two were gathered from the internet in order to be visually really different from the background images of the PRW dataset. When we use background images visually similar to what the network saw at training time, we observe that our approach is able to naturally integrate the foreground person into the corresponding background. Interestingly, the generated images  do not correspond to a simple pixel-to-pixel superimposition of the foreground image on top of the conditioning background. For instance, by comparing the third and fourth background columns, we see that the network adapts the brightness of the foreground to the brightness of the background. The image contrast and the blurring level in the generated images depend on the conditioning background. In the fifth column, second row, the network generated the bike wheel ahead of the person leg. A similar effect can be observed in the second background column, in which the legs of the people are partially occluded by the bikes. However, when we use images that are far from the background image distribution of the training set, our model fails to generate natural images. The backgrounds are correctly generated, but the persons are partially transparent.}

\newcommand\mySlash[2]{\ensuremath{%
  \!\sideset{_#1}{\!\!^#2}{\mathop\backslash}}}

\label{exp:BGcontrol}

\subsection{Ablation study and qualitative analysis}

\begin{figure}[h]
  \centering
  \setlength\tabcolsep{0.55pt}
\begin{tabular}{cccccccc}
$x_a$ & $P(x_a)$& $P(x_b)$& $x_b$  & \small\emph{Baseline}& \small\emph{DSC} & \small\emph{PercLoss} & \small\emph{Full}\\ 
\includegraphics[width=0.12\columnwidth]{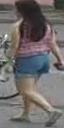}
&\includegraphics[width=0.12\columnwidth]{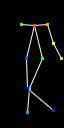} 
&\includegraphics[width=0.12\columnwidth]{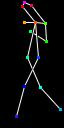}
&\includegraphics[width=0.12\columnwidth]{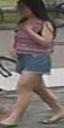}
&\includegraphics[width=0.12\columnwidth]{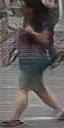}
&\includegraphics[width=0.12\columnwidth]{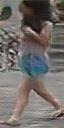}
&\includegraphics[width=0.12\columnwidth]{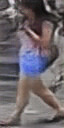}
&\includegraphics[width=0.12\columnwidth]{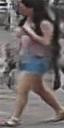}
\\ 
\includegraphics[width=0.12\columnwidth]{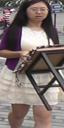}
&\includegraphics[width=0.12\columnwidth]{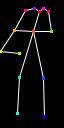} 
&\includegraphics[width=0.12\columnwidth]{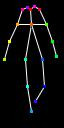}
&\includegraphics[width=0.12\columnwidth]{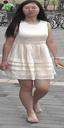}
&\includegraphics[width=0.12\columnwidth]{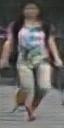}
&\includegraphics[width=0.12\columnwidth]{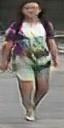}
&\includegraphics[width=0.12\columnwidth]{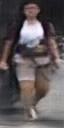}
&\includegraphics[width=0.12\columnwidth]{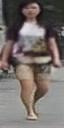}
\\ 
\includegraphics[width=0.12\columnwidth]{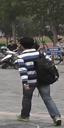}
&\includegraphics[width=0.12\columnwidth]{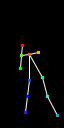} 
&\includegraphics[width=0.12\columnwidth]{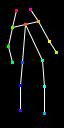}
&\includegraphics[width=0.12\columnwidth]{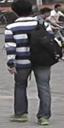}
&\includegraphics[width=0.12\columnwidth]{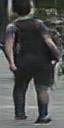}
&\includegraphics[width=0.12\columnwidth]{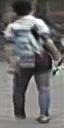}
&\includegraphics[width=0.12\columnwidth]{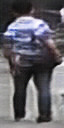}
&\includegraphics[width=0.12\columnwidth]{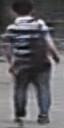}
\\ 
\includegraphics[width=0.12\columnwidth]{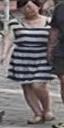}
&\includegraphics[width=0.12\columnwidth]{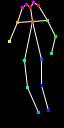} 
&\includegraphics[width=0.12\columnwidth]{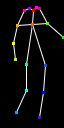}
&\includegraphics[width=0.12\columnwidth]{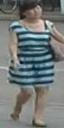}
&\includegraphics[width=0.12\columnwidth]{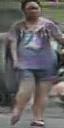}
&\includegraphics[width=0.12\columnwidth]{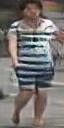}
&\includegraphics[width=0.12\columnwidth]{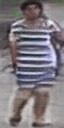}
&\includegraphics[width=0.12\columnwidth]{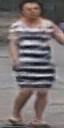}
\\ 
\includegraphics[width=0.12\columnwidth]{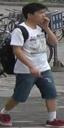}
&\includegraphics[width=0.12\columnwidth]{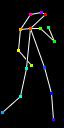} 
&\includegraphics[width=0.12\columnwidth]{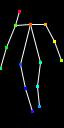}
&\includegraphics[width=0.12\columnwidth]{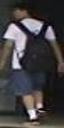}
&\includegraphics[width=0.12\columnwidth]{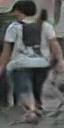}
&\includegraphics[width=0.12\columnwidth]{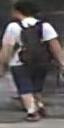}
&\includegraphics[width=0.12\columnwidth]{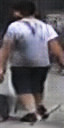}
&\includegraphics[width=0.12\columnwidth]{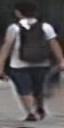}
\end{tabular}
  \caption{Qualitative results on the Market-1501 dataset. Columns 1, 2 and 3 represent the input of our model.
  We plot $P(\cdot)$  as a skeleton for the sake of clarity, but actually no joint-connectivity relation is exploited in our approach.
  Column 4 corresponds to the ground truth. The last four columns show the output of our approach with respect to different variants of our method.}
\label{fig:ablationMarket}
\end{figure}
\begin{figure}[h]
  \centering
  \setlength\tabcolsep{0.55pt}
\begin{tabular}{cccccccc}
$x_a$ & $P(x_a)$& $P(x_b)$& $x_b$  & \small\emph{Baseline}& \small\emph{DSC} &  \small\emph{PercLoss} & \small\emph{Full}\\ 
\includegraphics[width=0.12\columnwidth]{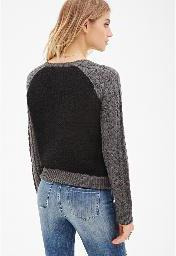}
&\includegraphics[width=0.12\columnwidth]{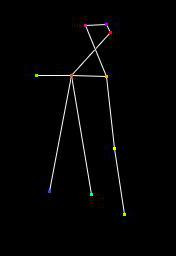} 
&\includegraphics[width=0.12\columnwidth]{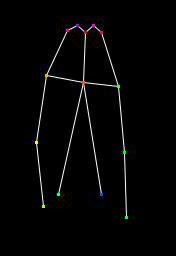}
&\includegraphics[width=0.12\columnwidth]{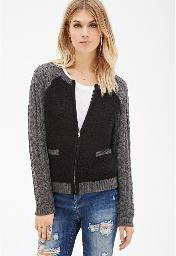}
&\includegraphics[width=0.12\columnwidth]{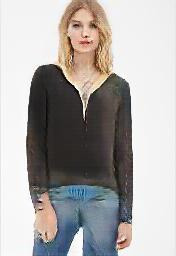}
&\includegraphics[width=0.12\columnwidth]{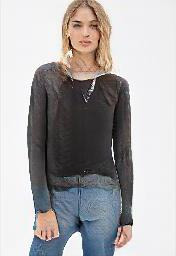}
&\includegraphics[width=0.12\columnwidth]{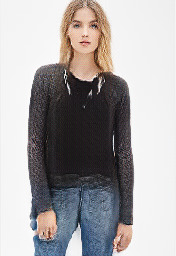}
&\includegraphics[width=0.12\columnwidth]{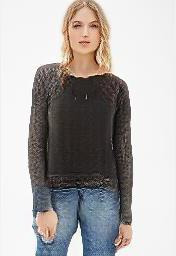}
\\ 
\includegraphics[width=0.12\columnwidth]{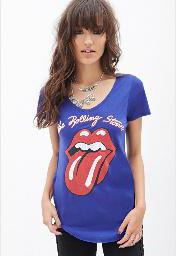}
&\includegraphics[width=0.12\columnwidth]{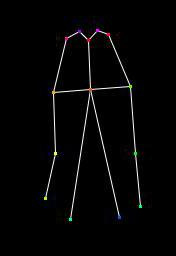} 
&\includegraphics[width=0.12\columnwidth]{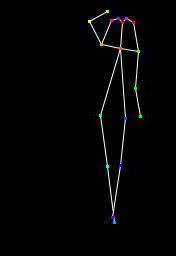}
&\includegraphics[width=0.12\columnwidth]{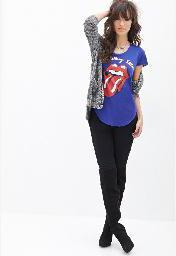}
&\includegraphics[width=0.12\columnwidth]{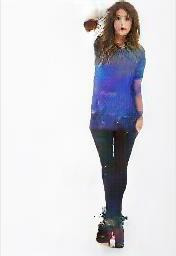}
&\includegraphics[width=0.12\columnwidth]{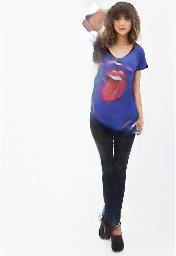}
&\includegraphics[width=0.12\columnwidth]{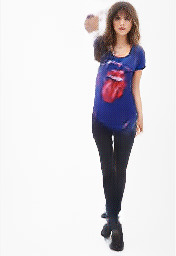}
&\includegraphics[width=0.12\columnwidth]{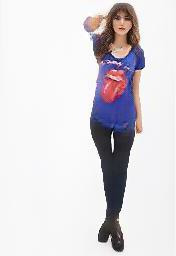}
\\ 
\includegraphics[width=0.12\columnwidth]{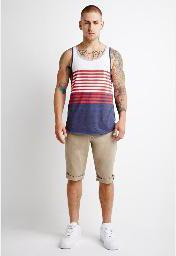}
&\includegraphics[width=0.12\columnwidth]{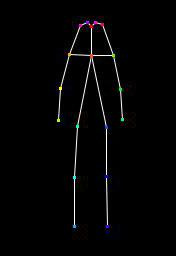} 
&\includegraphics[width=0.12\columnwidth]{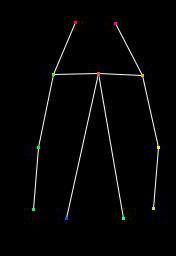}
&\includegraphics[width=0.12\columnwidth]{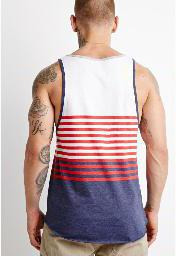}
&\includegraphics[width=0.12\columnwidth]{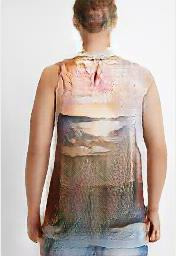}
&\includegraphics[width=0.12\columnwidth]{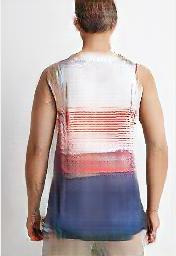}
&\includegraphics[width=0.12\columnwidth]{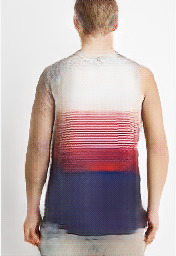}
&\includegraphics[width=0.12\columnwidth]{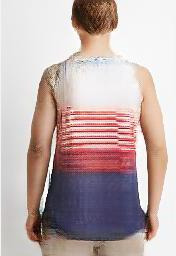}
\\ 
\includegraphics[width=0.12\columnwidth]{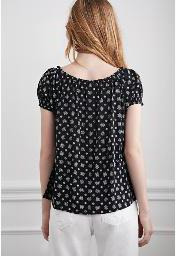}
&\includegraphics[width=0.12\columnwidth]{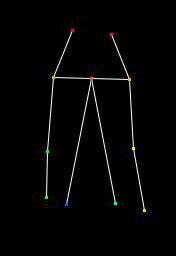} 
&\includegraphics[width=0.12\columnwidth]{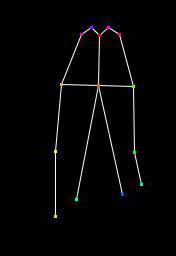}
&\includegraphics[width=0.12\columnwidth]{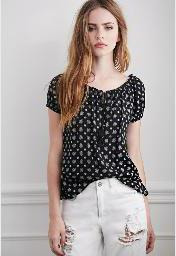}
&\includegraphics[width=0.12\columnwidth]{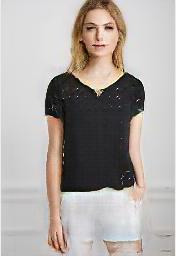}
&\includegraphics[width=0.12\columnwidth]{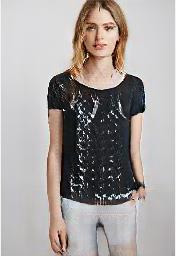}
&\includegraphics[width=0.12\columnwidth]{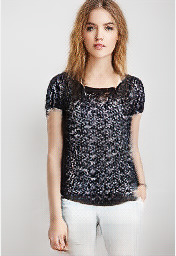}
&\includegraphics[width=0.12\columnwidth]{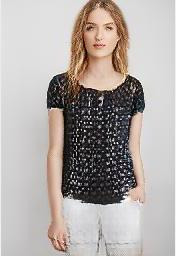}
\\
\includegraphics[width=0.12\columnwidth]{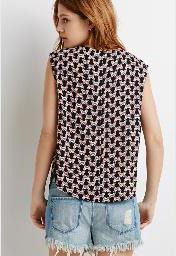}
&\includegraphics[width=0.12\columnwidth]{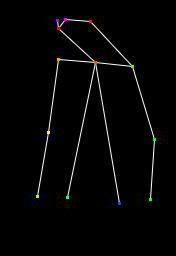} 
&\includegraphics[width=0.12\columnwidth]{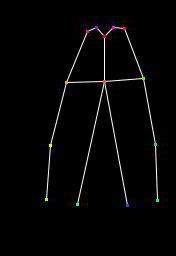}
&\includegraphics[width=0.12\columnwidth]{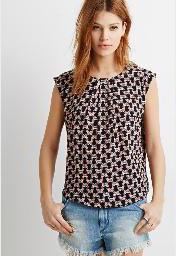}
&\includegraphics[width=0.12\columnwidth]{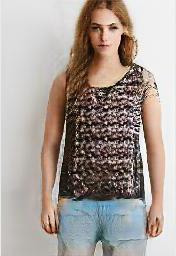}
&\includegraphics[width=0.12\columnwidth]{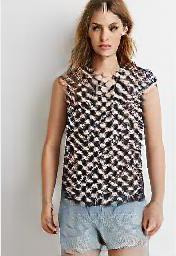}
&\includegraphics[width=0.12\columnwidth]{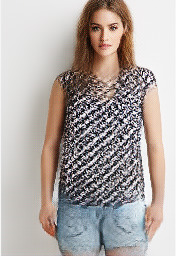}
&\includegraphics[width=0.12\columnwidth]{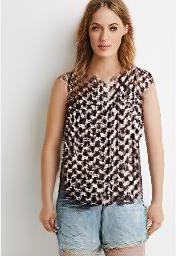}
\\
\includegraphics[width=0.12\columnwidth]{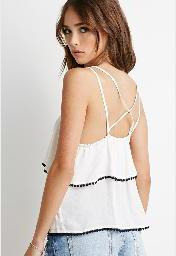}
&\includegraphics[width=0.12\columnwidth]{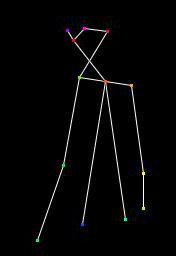} 
&\includegraphics[width=0.12\columnwidth]{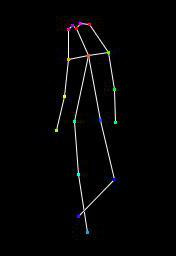}
&\includegraphics[width=0.12\columnwidth]{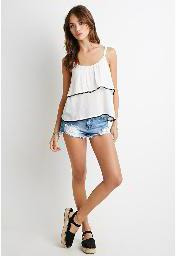}
&\includegraphics[width=0.12\columnwidth]{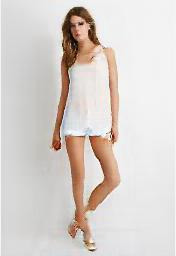}
&\includegraphics[width=0.12\columnwidth]{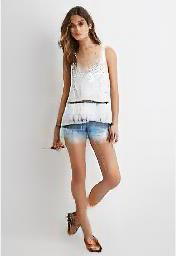}
&\includegraphics[width=0.12\columnwidth]{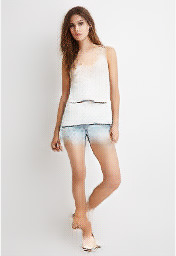}
&\includegraphics[width=0.12\columnwidth]{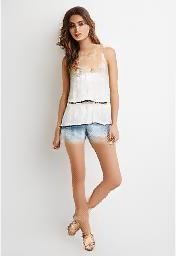}
\end{tabular}
  \caption{Qualitative results on the DeepFashion dataset with respect to different variants of our method. 
  Some images  have been cropped  to improve the visualization.}
\label{fig:ablationFashion}
\end{figure}

\begin{table*}[h]
\caption{Quantitative ablation study on the Market-1501 and the DeepFashion dataset.}
\centering
\begin{tabular}{l|ccccc|cc}
  \toprule
  &\multicolumn{5}{c|}{Market-1501}&\multicolumn{2}{c}{DeepFashion}\\
\midrule
  Model &\emph{SSIM} & \emph{IS}&\emph{mask-SSIM} & \emph{mask-IS} &\emph{DS}&\emph{SSIM} & \emph{IS}\\
\midrule
\emph{Baseline}&$0.256$ & $3.188$ & $0.784$ & $3.580$& $0.595$ &$0.754$ & $3.351$  \\
\emph{DSC}&$0.272$ & $\bf3.442$ & $0.796$ & $3.666$& $0.629$ &$0.754$ & $3.352$  \\
\emph{PercLoss} & $0.276$ & $3.342$ & $0.788$ & $3.519$ & $0.603$ & $0.744$ & $3.271$  \\
\emph{Full}&$\bf0.290$ & $3.185$ & $\bf0.805$ & $3.502$ & $\bf0.720$ &$\bf0.756$ & $\bf3.439$ \\
\midrule
\emph{Real-Data}&$1.00$ & $3.86$ & $1.00$ & \comRev{$\bf3.76$}& $0.74$ &$1.000$ & $3.898$ \\
\bottomrule
\end{tabular}
\label{tab:ablation}
\end{table*}

\label{Ablation}
In this section we present  an ablation study to clarify the impact of each part of our proposal on the final performance. We first describe the compared methods, obtained by ``amputating'' important parts of the full-pipeline presented in Sec.~\ref{architectures}. The discriminator architecture is the same for all the methods. 

\begin{itemize}
\item \emph{Baseline}:  We use the standard U-Net architecture  \cite{pix2pix2016} {\em without} deformable skip connections. The inputs of $G$ and $D$ and the way pose information is represented (see the definition of tensor $H$ in Sec.~\ref{architectures}) is the same as in 
the full-pipeline. However, in $G$, $x_a$, $H_a$ and $H_b$ are
 concatenated at the input layer. Hence,
 the encoder of $G$ is  composed of only one stream, whose architecture is the same as the two streams  described in Sec.~\ref{sec:Network-details}. 

\item \emph{DSC}: $G$ is implemented as described in Sec.~\ref{architectures}, introducing our Deformable Skip Connections (DSC). Both in DSC and in Baseline, training is performed using an $L_1$ loss together with the adversarial loss.

\item
\emph{PercLoss}: This is DSC in which the $L_1$ loss is replaced with the Perceptual loss proposed in \cite{DBLP:conf/eccv/JohnsonAF16}. 
This loss is computed using the layer $conv_{2\_1}$ of 
\cite{DBLP:journals/corr/SimonyanZ14a}, chosen to have a receptive field the closest possible to ${\cal N}( \mathbf{p})$ in Eq.~\ref{eq.L-NN-conv}, and computing the element-to-element difference in this layer  {\em without} nearest neighbour search.

\item \emph{Full}: This is the full-pipeline whose results are reported in 
Tabs.~\ref{tab:result}-\ref{tab:Re-ID}, and in which we use the proposed ${\cal L}_{NN}$ loss (see Sec.~\ref{Training}).
\end{itemize}

In Tab.~\ref{tab:ablation} we report a quantitative evaluation on the Market-1501 and on the DeepFashion dataset with respect to the four different versions of our approach. 
In most of the cases, there is a   progressive improvement  from  Baseline to  DSC to Full. Moreover, Full usually obtains better results than 
PercLoss. These improvements are particularly evident looking at the DS metrics.
 DS values on the DeepFashion dataset are omitted because they are all close to the value $\sim0.96$.

In Fig.~\ref{fig:ablationMarket} and Fig.~\ref{fig:ablationFashion} we show some qualitative results.
These figures show the progressive improvement through the four baselines which is quantitatively presented above.
In fact, while pose information is usually well generated by all the methods, 
the texture generated by Baseline often does not correspond to the texture in $x_a$ or is blurred. In same cases, the improvement of Full with respect to Baseline is quite drastic, such as the drawing on the shirt of the girl in the second row of Fig.~\ref{fig:ablationFashion} or 
the stripes on the clothes of the persons in the third and in the fourth row of Fig.~\ref{fig:ablationMarket}.

\begin{figure}[h]
  \centering
  \setlength\tabcolsep{0.5pt}
\begin{tabular}{cccccccc}
  $x_a$ & $P(x_a)$& $P(x_b)$& $x_b$  & \small\emph{Baseline }& \small\emph{DSC } & \small\emph{PercLoss } & \small\emph{Full }\\
  \includegraphics[width=0.11\columnwidth]{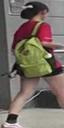}
&\includegraphics[width=0.11\columnwidth]{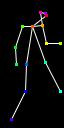} 
&\includegraphics[width=0.11\columnwidth]{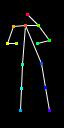}
&\includegraphics[width=0.11\columnwidth]{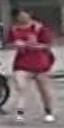}
&\includegraphics[width=0.11\columnwidth]{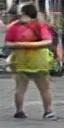}
&\includegraphics[width=0.11\columnwidth]{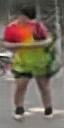}
&\includegraphics[width=0.11\columnwidth]{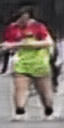}
&\includegraphics[width=0.11\columnwidth]{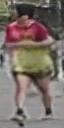}
\\
\includegraphics[width=0.11\columnwidth]{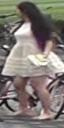}
&\includegraphics[width=0.11\columnwidth]{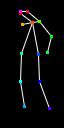} 
&\includegraphics[width=0.11\columnwidth]{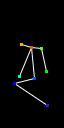}
&\includegraphics[width=0.11\columnwidth]{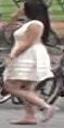}
&\includegraphics[width=0.11\columnwidth]{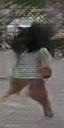}
&\includegraphics[width=0.11\columnwidth]{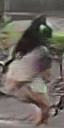}
&\includegraphics[width=0.11\columnwidth]{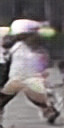}
&\includegraphics[width=0.11\columnwidth]{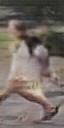}
\\
\includegraphics[width=0.11\columnwidth]{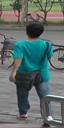}
&\includegraphics[width=0.11\columnwidth]{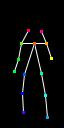} 
&\includegraphics[width=0.11\columnwidth]{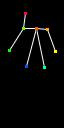}
&\includegraphics[width=0.11\columnwidth]{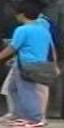}
&\includegraphics[width=0.11\columnwidth]{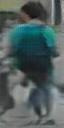}
&\includegraphics[width=0.11\columnwidth]{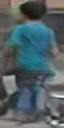}
&\includegraphics[width=0.11\columnwidth]{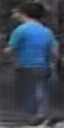}
&\includegraphics[width=0.11\columnwidth]{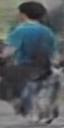}
\\
\includegraphics[width=0.11\columnwidth]{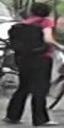}
&\includegraphics[width=0.11\columnwidth]{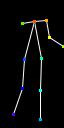} 
&\includegraphics[width=0.11\columnwidth]{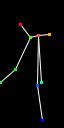}
&\includegraphics[width=0.11\columnwidth]{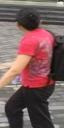}
&\includegraphics[width=0.11\columnwidth]{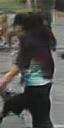}
&\includegraphics[width=0.11\columnwidth]{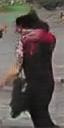}
&\includegraphics[width=0.11\columnwidth]{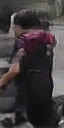}
&\includegraphics[width=0.11\columnwidth]{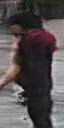}
\\
\includegraphics[width=0.11\columnwidth]{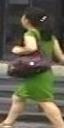}
&\includegraphics[width=0.11\columnwidth]{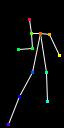} 
&\includegraphics[width=0.11\columnwidth]{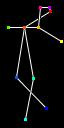}
&\includegraphics[width=0.11\columnwidth]{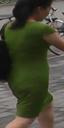}
&\includegraphics[width=0.11\columnwidth]{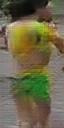}
&\includegraphics[width=0.11\columnwidth]{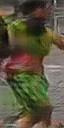}
&\includegraphics[width=0.11\columnwidth]{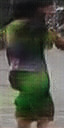}
&\includegraphics[width=0.11\columnwidth]{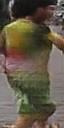}
\end{tabular}
  \caption{Examples of {\em badly} generated images on the Market-1501 dataset. See the text for more details.}
\label{fig:ablationMarket-Fail}
\end{figure}

\begin{figure}[h]
  \centering
  \setlength\tabcolsep{0.5pt}
\begin{tabular}{cccccccc}
  $x_a$ & $P(x_a)$& $P(x_b)$& $x_b$  & \small\emph{Baseline }& \small\emph{DSC } & \small\emph{PercLoss } & \small\emph{Full }\\
  \includegraphics[width=0.12\columnwidth]{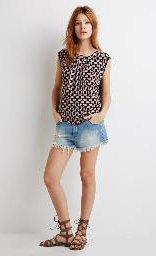}
&\includegraphics[width=0.12\columnwidth]{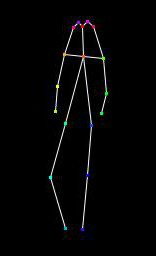} 
&\includegraphics[width=0.12\columnwidth]{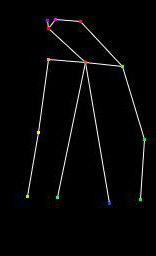}
&\includegraphics[width=0.12\columnwidth]{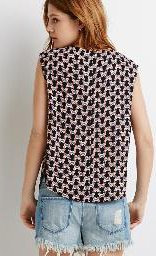}
&\includegraphics[width=0.12\columnwidth]{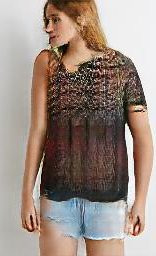}
&\includegraphics[width=0.12\columnwidth]{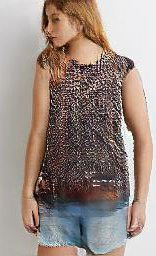}
&\includegraphics[width=0.12\columnwidth]{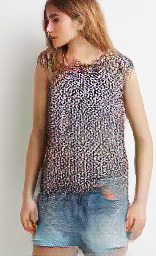}
&\includegraphics[width=0.12\columnwidth]{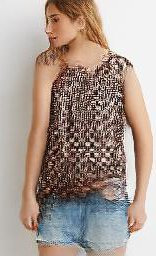}
\\
\includegraphics[width=0.12\columnwidth]{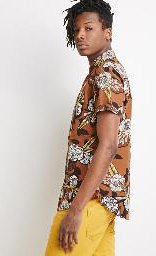}
&\includegraphics[width=0.12\columnwidth]{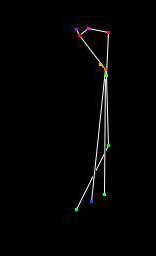} 
&\includegraphics[width=0.12\columnwidth]{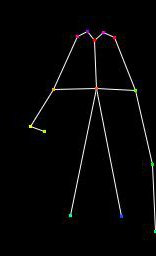}
&\includegraphics[width=0.12\columnwidth]{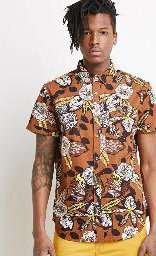}
&\includegraphics[width=0.12\columnwidth]{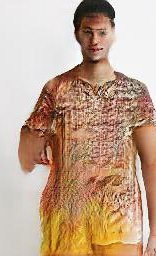}
&\includegraphics[width=0.12\columnwidth]{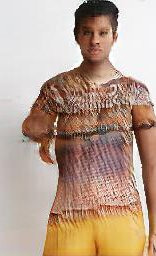}
&\includegraphics[width=0.12\columnwidth]{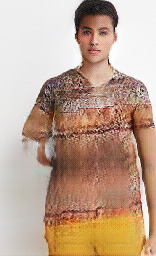}
&\includegraphics[width=0.12\columnwidth]{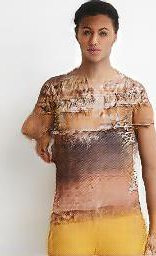}
\\
\includegraphics[width=0.12\columnwidth]{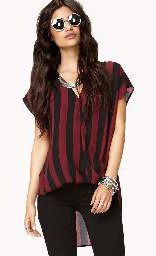}
&\includegraphics[width=0.12\columnwidth]{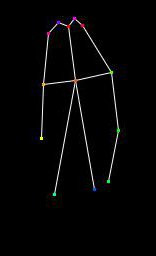} 
&\includegraphics[width=0.12\columnwidth]{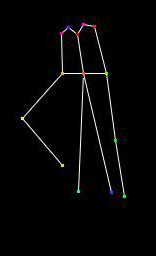}
&\includegraphics[width=0.12\columnwidth]{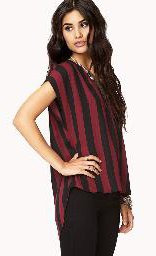}
&\includegraphics[width=0.12\columnwidth]{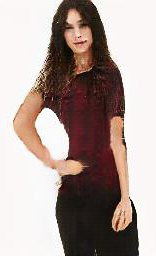}
&\includegraphics[width=0.12\columnwidth]{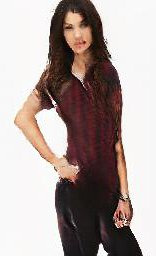}
&\includegraphics[width=0.12\columnwidth]{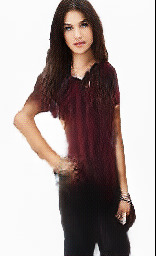}
&\includegraphics[width=0.12\columnwidth]{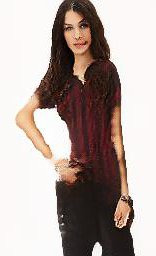}
\\
\includegraphics[width=0.12\columnwidth]{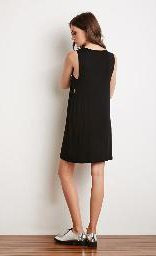}
&\includegraphics[width=0.12\columnwidth]{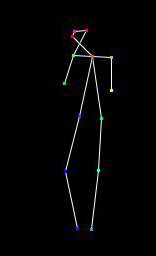} 
&\includegraphics[width=0.12\columnwidth]{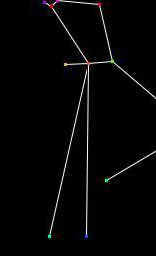}
&\includegraphics[width=0.12\columnwidth]{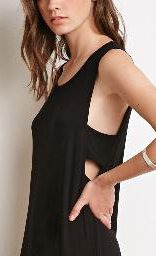}
&\includegraphics[width=0.12\columnwidth]{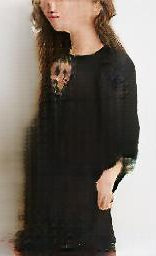}
&\includegraphics[width=0.12\columnwidth]{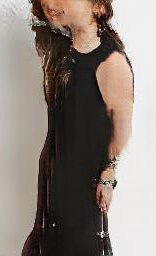}
&\includegraphics[width=0.12\columnwidth]{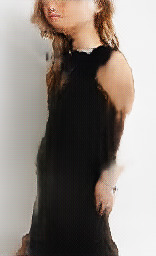}
&\includegraphics[width=0.12\columnwidth]{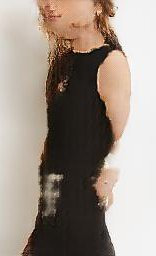}
\\
\includegraphics[width=0.12\columnwidth]{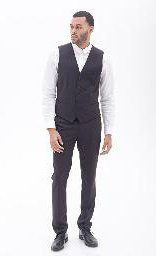}
&\includegraphics[width=0.12\columnwidth]{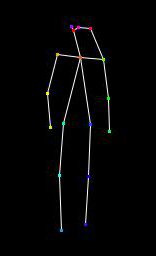} 
&\includegraphics[width=0.12\columnwidth]{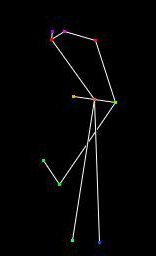}
&\includegraphics[width=0.12\columnwidth]{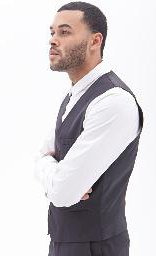}
&\includegraphics[width=0.12\columnwidth]{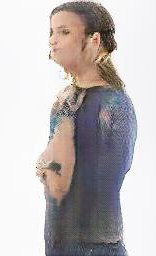}
&\includegraphics[width=0.12\columnwidth]{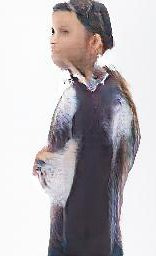}
&\includegraphics[width=0.12\columnwidth]{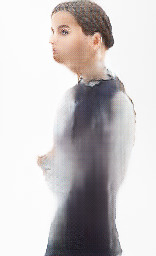}
&\includegraphics[width=0.12\columnwidth]{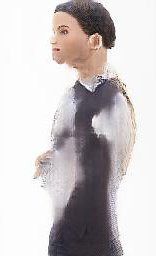}
\end{tabular}
  \caption{Examples of {\em badly} generated images on the DeepFashion dataset.}
\label{fig:ablationFashion-Fail}
\end{figure}

Finally,  Fig.~\ref{fig:ablationMarket-Fail} and 
Fig.~\ref{fig:ablationFashion-Fail} show some failure cases (badly generated images) of our method on the Market-1501 dataset and the DeepFashion dataset, respectively.
Some common failure  causes are:

\begin{itemize}
\item 
Errors of the HPE \cite{Cao}. For instance, see rows 2, 3 and 4 of  Fig.~\ref{fig:ablationMarket-Fail} or the wrong right-arm localization in row 2 of Fig.~\ref{fig:ablationFashion-Fail}.
\item
Ambiguity of the pose representation. 
For instance,
in  row 3 of Fig.~\ref{fig:ablationFashion-Fail},  the left elbow has been detected in $x_b$ although it is actually hidden behind the body. Since $P(x_b)$ contains only 2D information (no depth or occlusion-related information), there is no way for the system to understand whether the elbow is behind or in front of the body.  In this case our model chose to generate an arm considering that the arm is in front of the body (which corresponds to the most frequent situation in the training dataset). 
 \item 
 Rare poses. For instance, row 1 of Fig.~\ref{fig:ablationFashion-Fail} shows a girl in an unusual  rear view with a sharp 90 degree profile face ($x_b$). The generator by mistake synthesized a neck where it should have ``drawn'' a shoulder. Note that rare poses are a difficult issue also for other methods (e.g.,  \cite{ma2017pose}).
 \item 
 Rare object appearance. For instance, the backpack in row 1 of Fig.~\ref{fig:ablationMarket-Fail} is light green, while most of the backpacks contained in the training images of the Market-1501 dataset are dark. Comparing this image with the one generated in the last row of Fig.~\ref{fig:ablationMarket} (where the backpack is black), we see that in Fig.~\ref{fig:ablationMarket} the colour of the shirt of the generated image is not blended with the backpack colour, while in Fig.~\ref{fig:ablationMarket-Fail} it is.
 We presume  that the generator ``understands'' that a dark backpack is an object whose texture should not be transferred to the clothes of the generated image, while it is not able to generalize this knowledge to other backpacks. 
 \item 
 Warping problems. This is an issue related to our specific approach (the deformable skip connections). The texture on the shirt of the conditioning image in row 2 of Fig.~\ref{fig:ablationFashion-Fail} is warped in the generated image. We presume this is due to the fact that in this case the affine transformations  need to largely warp the texture details of the narrow surface of the profile shirt (conditioning image) in order to fit the much wider area of the target frontal pose.
\end{itemize}

\noindent
\addnote[maxCombined]{1}{\textbf{Combining the affine transformations.}
In our approach, we combine the different local affine transformations by selecting the maximum response as specified in Eq.~\eqref{eq.d-F}. Alternatively, we could use the average operator to combine the local affine transformations leading to a soft combination. A third approach could consist in combining the transformations via linear combination. In that case, the features at location $\mathbf{p}$, $F'_h(\mathbf{p},c)$, are combined following:
\begin{equation}
\label{eq.d-F-inter}
d(F(\mathbf{p},c)) = \sum_{h = 1, ..., 10}\omega_{h}(\mathbf{p})\odot F'_h(\mathbf{p},c),
\end{equation}
where  $\omega_{h}(\mathbf{p})\in\mathbb{R}$ are the weights of the linear combination. The intuition behind this formulation is that, at each location $\mathbf{p}$, we use the weights $\omega_h(\mathbf{p})$ to select in a soft manner the relevant features $F'_h(\mathbf{p},\cdot)$. In order to condition $\omega_{h}$ on the feature maps $F'_h$, we feed the concatenation of the $F'_h$ tensors (with $h=1,...,10$) along the channel axis into a 1x1 convolution layer that returns all the $\omega_{h}$ maps. We normalize $\omega_{h}$ in order to sum to one at each location $\mathbf{p}$.
We compare these two soft approaches with our max-based approach in Tab.\ref{tab:ablationCombi} on the Market-1501. These results show that both the average and the linear combination approaches perform well but slightly worst than the max-based formulation of Eq.\ref{eq.d-F-inter}. The performance differences are especially clear when looking at the mask-based metrics and the detection scores. In addition, we observe that the linear combination model under-performs both the \emph{average} and the \emph{maximum} approaches.
\begin{table}[h]
\caption{\comRev{Combining the affine transformations: quantitative ablation study on the Market-1501.}}
\centering
\begin{tabular}{l|ccccc}
  \toprule
  &\multicolumn{5}{c}{Market-1501}\\
\midrule
 Combination &\emph{SSIM} & \emph{IS}&\emph{mask-SSIM} & \emph{mask-IS} &\emph{DS}\\
\midrule
Average& $\bf 0.290$ & $\bf 3.217$ & $0.802$ & $3.428$ & $0.695$ \\
Linear Comb.& $0.287$&$3.078$  & $0.799$ & $3.385$ & $0.703$\\
Max& $\bf 0.290$ & $3.185$ & $\bf 0.805$ & $\bf 3.502$ & $\bf 0.720$\\
\bottomrule
\end{tabular}
\label{tab:ablationCombi}
\end{table}
}

\noindent
\addnote[expePose]{1}{\textbf{Robustness to HPE errors.}
We now evaluate the robustness of our model to HPE errors. We propose to perform this evaluation using the following protocol. First, we train a model in a standard way. Then, at test time, we simulate HPE errors by randomly perturbing the limb positions predicted by the HPE. More precisely, the source poses are randomly perturbed using an isotropic 0-mean Gaussian noise to the arm and the leg landmarks with standard deviations equal to $\sigma$ pixels. This experiment is conducted with standard deviations $\sigma$ varying from 0 to 25 pixels. We model HPE errors by adding noise to the arm and leg landmarks since this corresponds to the most frequent HPE errors. The performances are reported in Fig.~\ref{fig:noise} in term of mask-SSIM and mask-IS.
\begin{figure}[t]\centering
\includegraphics[width=0.99\columnwidth]{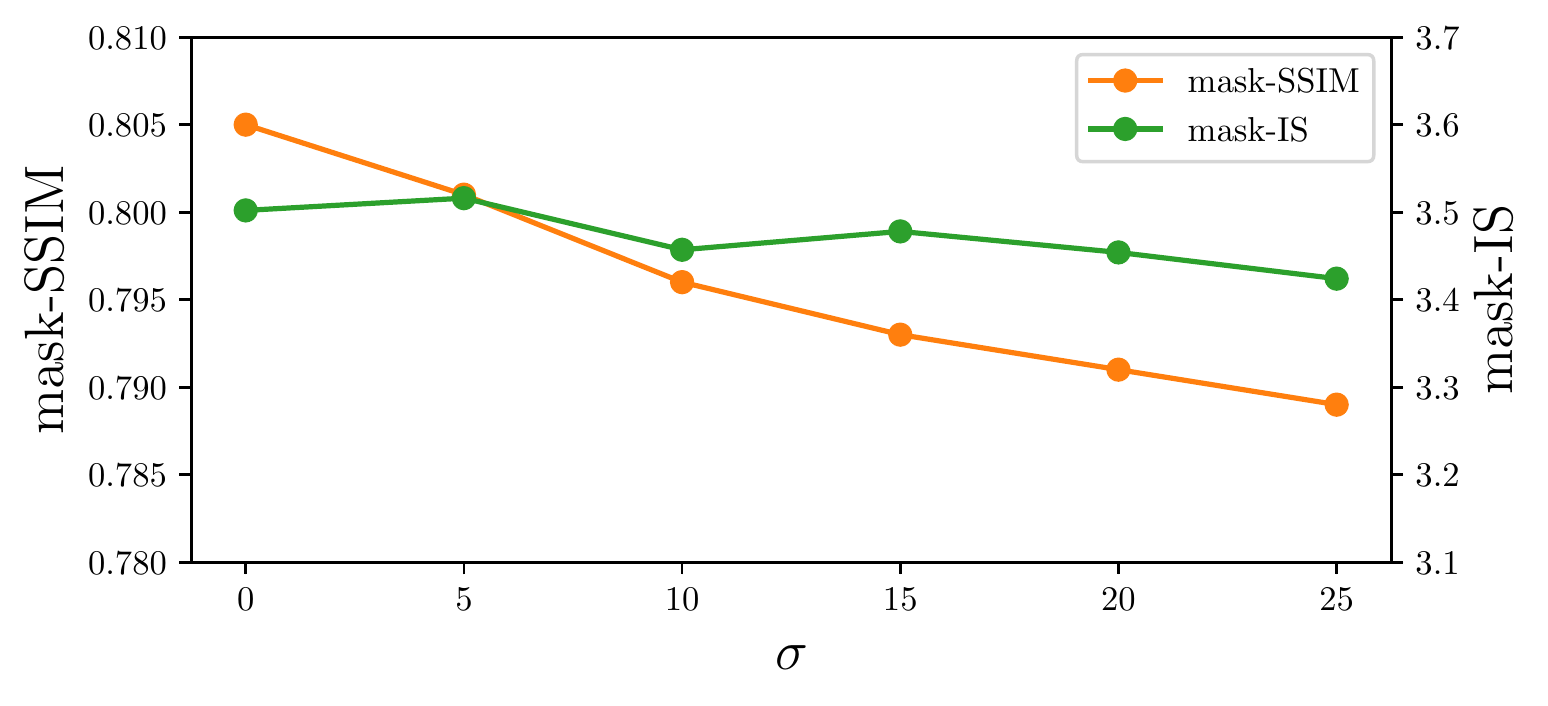}\vspace{-0.3cm}
\caption{\comRev{Robustness to HPE errors: at test time we randomly perturbing the estimated limb position with a Gaussiam noise with a standard deviation $\sigma$.}}
\vspace{-0.5cm}
\label{fig:noise}
\end{figure}
We observe that mask-SSIM decreases consistently when the HPE error increases. Since mask-SSIM measures the reconstruction quality, this shows that an accurate HPE helps to reconstruct well the details of the person in the input image. Conversely, we observe that mask-IS values are not much affected by the HPE errors. This can be explained by the fact that IS-based metrics measure image quality and diversity but not reconstruction. This experiment shows that HPE errors impact the model ability to preserve the appearance of the conditioning image, but do not affect significantly the quality and diversity of the generated images. Nevertheless, even in the case of very noisy HPE, e.g., $\sigma=25$ pixels, our model performs similarly to the \emph{Baseline} model in Tab.~\ref{tab:ablation} which does not use deformable skip connections. This shows that our proposed deformable skip connections are robust to HPE errors.}

\noindent
\addnote[expeGfunc]{1}{\textbf{Choice of the $g$ function.}
Our nearest neighbour loss uses an auxiliary function $g$. In order to measure the impact of the choice of $g$, in Table \ref{tab:ablationBlock}, we compare the scores obtained when using different layers of the VGG-19 network to implement $g$. We observe that when our nearest neighbour loss is computed in the pixel space, we obtain good mask-SSIM and mask-IS scores but really poor detection scores. It shows that, assessing reconstruction quality in the pixel space leads to images with a poor general structure. Conversely, when we use a higher network layer to implement $g$, we obtain lower mask-SSIM and mask-IS scores but better detection score. Note that, using a higher layer for $g$ (\textit{i.e.,} Block $3\_1$) also increases the training computation. Finally, our proposed implementation that uses Block $1\_2$ for $g$, reaches the highest detection score and the best trade-off between the other metrics.
\begin{table}[h]
\caption{\comRev{Choice of the $g$ function: quantitative ablation study on the Market-1501.}}
\centering
\begin{tabular}{l|ccccc}
  \toprule
  &\multicolumn{5}{c}{Market-1501}\\
\midrule
  $g$ function &\emph{SSIM} & \emph{IS}&\emph{mask-SSIM} & \emph{mask-IS} &\emph{DS}\\
\midrule
Pixel space& $0.285$ &$\bf 3.339$  & $\bf0.806$ & $\bf3.630$ & $0.580$\\
Block $1\_2$&$\bf0.290$ & $3.185$ & $0.805$ & $3.502$ & $\bf0.720$\\
Block $2\_1$& $0.283$ &$3.034$  & $0.802$ & $3.388$ & $0.648$\\
Block $3\_1$& $0.244$ &$3.319$  & $0.794$ & $3.187$ & $\bf0.699$\\
\bottomrule
\end{tabular}
\label{tab:ablationBlock}
\end{table}
}

\begin{figure}[h]
  \centering
  \setlength\tabcolsep{0.5pt}
  \begin{tabular}{ccccc}
   \multirow{2}{*}{$x_a$} &\multirow{2}{*}{$x_b$}&\multicolumn{3}{c}{$\lambda$ value}\\
   &   & $0.1$& $0.01$&$0.001$\\
  \includegraphics[width=0.11\columnwidth]{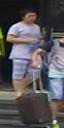}&
  \includegraphics[width=0.11\columnwidth]{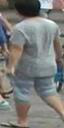}&
  \includegraphics[width=0.11\columnwidth]{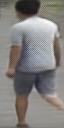}&
  \includegraphics[width=0.11\columnwidth]{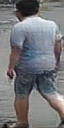}&
  \includegraphics[width=0.11\columnwidth]{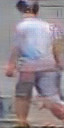}
  \\
  \includegraphics[width=0.11\columnwidth]{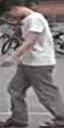}&
  \includegraphics[width=0.11\columnwidth]{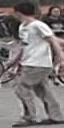}&
  \includegraphics[width=0.11\columnwidth]{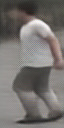}&
  \includegraphics[width=0.11\columnwidth]{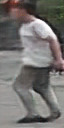}&
  \includegraphics[width=0.11\columnwidth]{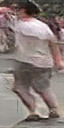}
  \\
    \includegraphics[width=0.11\columnwidth]{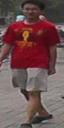}&
  \includegraphics[width=0.11\columnwidth]{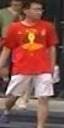}&
  \includegraphics[width=0.11\columnwidth]{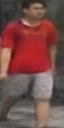}&
  \includegraphics[width=0.11\columnwidth]{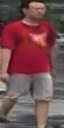}&
  \includegraphics[width=0.11\columnwidth]{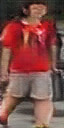}
  \\
    \includegraphics[width=0.11\columnwidth]{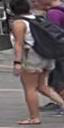}&
  \includegraphics[width=0.11\columnwidth]{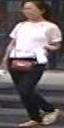}&
  \includegraphics[width=0.11\columnwidth]{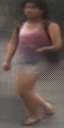}&
  \includegraphics[width=0.11\columnwidth]{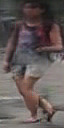}&
  \includegraphics[width=0.11\columnwidth]{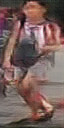}\\
\end{tabular}
  \caption{\comRev{Qualitative ablation study: impact of the $\lambda$ parameter.}}
\label{fig:ablationQualLambda}
\end{figure}

  \noindent
\addnote[expelambda]{1}{\textbf{Sensitivity to the $\lambda$ parameter.} As training objective, we use a combination of a reconstruction and an adversarial loss as specified in Eq.~\eqref{eq.objective}. We now evaluate the impact of the $\lambda$ parameter that controls the balance  between the two losses. Quantitative results are reported in Tab.~\ref{tab:ablationLam}. We observe that a low $\lambda$ value leads to better SSIM scores but lower IS values. Indeed, since SSIM measures reconstruction quality, a low $\lambda$ value reduces the impact of the adversarial loss and therefore generates images with a high reconstruction quality. Consequently, it also reduces diversity and realism leading to poorer IS values. Conversely, with a high $\lambda$ value, we obtain high IS values but low SSIM scores. We further investigate the impact of the $\lambda$ parameter by reporting a qualitative comparison in Fig~\ref{fig:ablationQualLambda}. Consistently with the quantitative comparison, we observe that lower $\lambda$ values result in smoother images without texture details whereas high values generate detailed images but with more artifacts.}
\begin{table}[h]
\caption{\comRev{Quantitative ablation study on the Market-1501: sensitivity to the $\lambda$ parameter}}
\centering
\begin{tabular}{l|ccccc}
  \toprule
  &\multicolumn{5}{c}{Market-1501}\\
\midrule
 $\lambda$ &\emph{SSIM} & \emph{IS}&\emph{mask-SSIM} & \emph{mask-IS} &\emph{DS}\\
\midrule
0.1& $\bf0.292$&$2.621$  & $\bf0.808$ & $3.168$ & $0.697$\\
0.01&$0.290$ & $3.185$ & $0.805$ & $3.502$ & $\bf0.720$\\
0.001& 0.245& \bf3.566  &0.779  & $\bf3.634$ & $0.609$\\
\bottomrule
\end{tabular}
\label{tab:ablationLam}
\end{table}

  \noindent
  \addnote[expeSym]{1}{\textbf{Exploiting symmetry.~}
  In Sec.~\ref{symmetry}, we explained how we exploit human-body symmetry to improve generation.
When a missing region involves a limb whose symmetric body part is detected, we copy the features from the
twin part. We now evaluate the impact of this strategy. For comparison, we introduce a model that does not use symmetry. More precisely, the
regions and the affine transformations corresponding to undetected body parts are not computed
and the corresponding region-masks are filled with 0. Results are reported in Table ~\ref{tab:ablationSym}. We observe that exploiting human-body symmetry with the proposed consistently improves all the metrics.
Although limited, these consistent gains clearly show the benefit of our approach.
  \begin{table}[h]
\caption{\comRev{Quantitative ablation study on the Market-1501: Exploiting symmetry.}}
\centering
\begin{tabular}{l|ccccc}
  \toprule
  &\multicolumn{5}{c}{Market-1501}\\
\midrule
 Symmetry &\emph{SSIM} & \emph{IS}&\emph{mask-SSIM} & \emph{mask-IS} &\emph{DS}\\
\midrule
Without& $0.289$&$3.130$  & $0.801$ & $3.464$ & $0.698$\\
With& $\bf 0.290$ & $\bf 3.185$ & $\bf 0.805$ & $\bf 3.502$ & $\bf 0.720$\\
\bottomrule
\end{tabular}
\label{tab:ablationSym}
\end{table}}

  \noindent
  \addnote[expeNN]{2}{\textbf{Neighborhood size.~}
In Sec.\ref{Training}, we introduce a nearest-neighbour loss to improve generation quality. This loss is defined as a minimum feature distance within a local neighborhood. The neighborhood size is defined in Eq.~\eqref{eq.L-NN} using a parameter $n$.
 We now evaluate the impact of this parameter. Results on the Market-1501 datasets are reported in Tab.~\ref{tab:ablationNN}. We also report the performance of the \emph{DSC} baselines that do not use our nearest-neighbour loss. First, we observe that the models with $n=1$ or $n=3$ obtain better \emph{Detection Score} than the \emph{DSC} model. This shows that the model is relatively robust to different neighborhood sizes. Nevertheless, the \emph{Detection Score} drops when we use a too large neighborhood size, e.g., $n=5$, and reaches a lower score than \emph{DSC}. This shows that, when the reconstruction loss tolerates too large misalignments, it affects the performance.  
Second, we observe that low n values lead to better IS scores but lower SSIM values. Especially, the model with $n=3$ returns a significantly higher \emph{mask-SSIM} value than when $n=1$. This shows that tolerating small spatial misalignments at training time helps image reconstruction.
Finally, concerning the DeepFashion dataset, we increased $n$ to $n=5$ in order to cope with the higher resolution of the images of this benchmark.
  \begin{table}[h]
\caption{\comRevSec{Quantitative ablation study on the Market-1501: neighborhood size.}}
\centering
\begin{tabular}{>{\color{darkgreen}}l|>{\color{darkgreen}}c>{\color{darkgreen}}c>{\color{darkgreen}}c>{\color{darkgreen}}c>{\color{darkgreen}}c}
  \toprule
  &\multicolumn{5}{>{\color{darkgreen}}c}{Market-1501}\\
\midrule
 Model &\emph{SSIM} & \emph{IS}&\emph{mask-SSIM} & \emph{mask-IS} &\emph{DS}\\
\midrule
$n=1$& $0.289$ & $\bf3.398$ & $0.755$ & $\bf3.553$ & $\bf0.728$\\
$n=3$& $\bf 0.290$ & $ 3.185$ & $\bf 0.805$ & $3.502$ & $0.720$\\
$n=5$& $0.289$ & $3.144$ & $0.802$ & $3.487$ & $0.579$\\
\midrule
\emph{DSC}&$0.272$ & $\bf 3.442$ & $0.796$ & $\bf 3.666$& $0.629$\\
\bottomrule
\end{tabular}
\label{tab:ablationNN}
\end{table}}

 \noindent
  \addnote[expeMask]{2}{\textbf{Importance of Mask.~}
In Sec.~\ref{skip-connections}, we detail how we align the feature of each body-part separately. In particular, in Eq.~\eqref{eq.F}, we mask out the irrelevent regions before warping the feature maps. We now evaluate the impact of using masks. We compare our model with a model without masks ($F'_h = f_h (F)$ in \eqref{eq.F}). Results on the Market-1501 datasets are reported in Table ~\ref{tab:ablationMask}.
 We observe that considering masks when aligning features improves all the metrics. In particular, it improves the \emph{DS} value by $3\%$. These consistent gains show the benefit of our masked-based formulation.
  \begin{table}[h]
\caption{\comRevSec{Quantitative ablation study on the Market-1501: impact of using mask.}}
\centering
\begin{tabular}{>{\color{darkgreen}}l|>{\color{darkgreen}}c>{\color{darkgreen}}c>{\color{darkgreen}}c>{\color{darkgreen}}c>{\color{darkgreen}}c}
  \toprule
  &\multicolumn{5}{>{\color{darkgreen}}c}{Market-1501}\\
\midrule
 Mask &\emph{SSIM} & \emph{IS}&\emph{mask-SSIM} & \emph{mask-IS} &\emph{DS}\\
\midrule
$Without$& $0.286$ & $3.048$ & $0.800$ & $3.4388$ & $0.693$\\
$With$& $\bf 0.290$ & $ \bf 3.185$ & $\bf 0.805$ & $\bf 3.502$ & $\bf0.720$\\
\bottomrule
\end{tabular}
\label{tab:ablationMask}
\end{table}}

%% file: exp-RID.tex
\subsection{Person generation for Re-ID data-augmentation}
\label{PersonRe-ID}

The experiments in this section are
motivated by the importance of using generative methods as a data-augmentation tool which provides additional labeled samples for training discriminative methods (see Sec.~\ref{Introduction}).
Specifically, we show here that the synthetic images generated by our Deformable GANs can be used to train different Re-ID networks. The typical Re-ID  task consists in recognizing the identity of a human person in different poses, viewpoints and scenes. The common  application  of a Re-ID system is a video-surveillance  scenario  in which
 images of the same person, grabbed by 
cameras mounted in different  locations, need to be matched to  each other. Due to the low-resolution of the cameras, person Re-ID is usually based on the colours and the texture of the clothes  \cite{DBLP:journals/corr/ZhengYH16}. This makes our method particularly suited to automatically populate a Re-ID training dataset by generating images of a given person with identical clothes but in different viewpoints/poses. 

In our   experiments we use   different 
Re-ID methods, taken from \cite{DBLP:journals/corr/ZhengYH16,DBLP:journals/tomccap/ZhengZY18}. \addnote[expeRID]{1}{First, IDE~\cite{DBLP:journals/corr/ZhengYH16} is an approach that consists in regarding Re-ID training as an image classification task where each class corresponds to a person {\em identity}. At test time, the identity is assigned based on the image feature representation obtained before the classification layer of the network. Each query image is associated to the {\em identity} of the closest image in the gallery. Different metrics can be employed at this stage to determine the closest gallery image. In our experiments, we consider three metrics: the Euclidean distance, a metric based on Cross-view Quadratic Discriminant Analysis (XQDA~\cite{liao2015person}) and a Mahalanobis-based distance (KISSME \cite{koestinger2012large}). Second, opposed to the IDE approach that predicts identity labels, in \cite{DBLP:journals/tomccap/ZhengZY18}, a siamese network predicts whether the identities of the two input images are the same. For all approaches, we use a ResNet-50 backbone pre-trained on ImageNet. Here, these Re-ID methods are used as black-boxes and trained with or without data-augmentation. We refer the reader to the corresponding articles for additional details about the involved approaches.}

For training and testing we use the Market-1501 dataset that is designed for Re-ID  benchmarking.
Since \cite{esser2018variational,ma2018disentangled} cannot be explicitly conditioned on a background
image, for a fair comparison we also tested our Deformable GANs {\em without} background conditioning (Sec.~\ref{subSec:BGcontrol}).
For each of the tested person-generation approaches, we use the following data augmentation procedure. 
In order to augment the Market-1501 training dataset ($T$ of size $|T|$) by a factor $\alpha$,
for each image in $T$ we randomly select $\tau = \alpha - 1$ target poses, generating $\tau$ corresponding images using a person-generation approach.
Note that: (1) Each generated image is labeled with the {\em identity} of the conditioning appearance image, (2) 
The target {\em pose} can be extracted from an individual different from the person depicted in the conditioning appearance image.
Adding the generated $\tau |T|$ images to $T$ we obtain an augmented training set. 

In Tab.~\ref{tab:Re-ID} 
we report the results obtained using either $T$ (standard procedure, $\tau = 0$) or the augmented dataset for training different Re-ID systems. 
Each row of the table corresponds to a different generative method used for data augmentation. Specifically, the results corresponding to our Deformable GANs are presented using two variants of our method: the {\em Full} pipeline, as described in Sec.~\ref{architectures} and using a {\em Baseline} generator architecture with an heat-map based pose representation (Eq.~\ref{eq.blurring}) but without deformable skip connections and nearest-neighbour loss (see Sec.~\ref{Ablation} for more details).

The other person-generation approaches used for data augmentation are: \cite{esser2018variational,ma2018disentangled,liu2018pose}. 
Note that in \cite{ma2018disentangled} the data-augmentation procedure is slightly different from  the one used by all the other methods (ours included). Indeed, in \cite{ma2018disentangled}, new person appearances are synthesized by sampling appearance descriptors in a preliminarly learned embedding. 
Moreover, \cite{liu2018pose} is the only method which is specifically designed and trained for Re-ID data augmentation, using a Re-ID based loss to specifically drive the person-generation task
 (see Sec.~\ref{Related}), while all the other tested approaches, including ours, generate images {\em independently} of the Re-ID task which is used in this section for testing.  
Liu et al. \cite{liu2018pose} report slightly better results ({\em Rank 1 = 79.7} and {\em mAP = 57.9}) obtained using Label Smoothing Regularization (LSR) \cite{DBLP:journals/corr/SzegedyVISW15}
when training {\em the final Re-ID system}. However, LSR is based on a confidence value 
used to weight the generated samples differently from the real samples, and this 
hyper-parameter needs to be manually tuned depending on how trustable the generator is.  
For a fair comparison with other methods which do not adopt LSR (including ours), 
in Tab.~\ref{tab:Re-ID}, column ``IDE + Euclidean'',
we report the 
results obtained 
by \cite{liu2018pose} without LSR when
training the final Re-ID system.

Tab.~\ref{tab:Re-ID} shows a significant accuracy boost when using our full model with respect to using only $T$.
 This dramatic performance boost, orthogonal to different Re-ID methods, 
shows that our generative approach can be effectively used 
 for synthesizing training samples.
It also indirectly shows that the generated images are sufficiently realistic and different from the real images contained in $T$. Importantly, we notice that there is no boost when using the Baseline model. Conversely, the Baseline-based results are even lower than without data-augmentation and 
this accuracy decrement is even more drastic when $\alpha = 10$ (higher data-augmentation factor).
The comparison between Full and Baseline  shows the importance of the proposed method  to get sufficiently realistic images. Interestingly, we observe a similar, significant {\em negative} accuracy difference when data-augmentation is performed using either  \cite{ma2018disentangled} or \cite{esser2018variational}. 
Our Full results are even 
slightly better 
than \cite{liu2018pose}, despite the latter method is specifically designed for person Re-ID and data are generated driven by a Re-ID loss during training.
These results indirectly
confirm that our Deformable GANs can effectively  capture  person-specific details
which are important to identify a person.

  \addnote[expeRID-10]{1}{Interestingly, we observe that with a high augmentation factor, such as $\alpha=10$, the performance is lower than with a factor 2. This indicates that increasing the number of generated images may harm Re-ID performance. This observation may seem counterintuitive since, in standard scenarios, the more data the better. A possible explanation for this drop in performance is that the proportion of real, and so artefact-free data is reduced when the augmentation factor increases.}

\begin{table*}[h]
\caption{Influence of person-generation based data augmentation on the accuracy of different Re-ID methods on the Market-1501 test set (\emph{Rank 1}~/~\emph{mAP} in $\%$). 
$(*)$ Uses a different data-augmentation strategy (see details in the text).
$(**)$ These results have been provided by the authors of \cite{liu2018pose} via personal communication and are slightly different from those reported in \cite{liu2018pose}, the latter being obtained by training the Re-ID network using LSR (see the text for more details).} 
\centering

\begin{tabular}{lccccccccc}
    \toprule
&Augmentation& \multicolumn{2}{c}{IDE + Euclidean \cite{DBLP:journals/corr/ZhengYH16}}    & \multicolumn{2}{c}{IDE + XQDA \cite{DBLP:journals/corr/ZhengYH16}} &\multicolumn{2}{c}{IDE + KISSME \cite{DBLP:journals/corr/ZhengYH16}} & \multicolumn{2}{c}{Discriminative Embedding \cite{DBLP:journals/tomccap/ZhengZY18}}  \\
&factor ($\alpha$) &\emph{Rank 1}&\emph{mAP} & \emph{Rank 1}&\emph{mAP}& \emph{Rank 1}&\emph{mAP}& \emph{Rank 1}&\emph{mAP}\\
  \midrule
    No augmentation ($T$) &1 & 73.9 & 48.8 &73.2 & 50.9 &75.1 & 51.5 & 78.3  & 55.5\\
\midrule
    Ma et al. $(*)$ \cite{ma2018disentangled}&$\approx{2}$ &66.9 & 41.7 &69.9 & 47.4 & 71.9 & 47.7 & 73.9  & 51.6\\
    Esser et al. \cite{esser2018variational}&2 &58.1 & 33.7 & 68.9 & 46.1 & 67.8 & 46.1 & 63.1 & 40.3 \\
    Liu et al.\cite{liu2018pose} $(**)$ & 2  & 77.9 
     & 56.62  
     & - &- &- &- &- &- \\

\midrule

{\em Ours (Baseline)} & 2 &68.1 & 42.82 & 69.57  & 46.43 & 69.45 & 45.88 & 70.69 &46.58  \\
{\em Ours (Full)} & 2 & \bf 78.9 & \bf 56.9 & \bf 78.2 & \bf57.9  & \bf 79.7 & \bf 58.3 & \bf 81.4 & 60.3 \\  

\midrule
{\em Ours (Baseline)} & 10 &59.8 & 34.5 & 60.9 & 38.2 & 61.9 & 37.8 & 61.6 & 39.4 \\

{\em Ours (Full)} & 10 &  78.5  &  55.9 & 77.8  & \bf 57.9  & 79.5  & 58.1 & 80.6  &  \bf 61.3  \\

\bottomrule
\end{tabular}

\label{tab:Re-ID}
\end{table*}